\documentclass[10pt,journal,compsoc]{IEEEtran}
\ifCLASSOPTIONcompsoc
  \usepackage[nocompress]{cite}
\else
  \usepackage{cite}
\fi
\ifCLASSINFOpdf
\else
\fi
\usepackage{epsfig}
\usepackage{graphicx}
\usepackage{booktabs}       
\usepackage{amsfonts}       
\usepackage{amsthm}
\usepackage{nicefrac}       
\usepackage{microtype}      
\usepackage{xcolor,pifont}
\usepackage{epsfig}
\usepackage{subfigure}
\usepackage{amsmath}
\usepackage{array}
\usepackage{pifont}
\usepackage{wrapfig}
\usepackage{hhline}
\usepackage{mathrsfs} 
\usepackage{algorithm}
\usepackage{listings}
\usepackage{xcolor,amsmath}

\usepackage{etoolbox}
\makeatletter
\AfterEndEnvironment{algorithm}{\let\@algcomment\relax}
\AtEndEnvironment{algorithm}{\kern2pt\hrule\relax\vskip3pt\@algcomment}
\let\@algcomment\relax
\newcommand\algcomment[1]{\def\@algcomment{\footnotesize#1}}
\renewcommand\fs@ruled{\def\@fs@cfont{\bfseries}\let\@fs@capt\floatc@ruled
  \def\@fs@pre{\hrule height.8pt depth0pt \kern2pt}%
  \def\@fs@post{}%
  \def\@fs@mid{\kern2pt\hrule\kern2pt}%
  \let\@fs@iftopcapt\iftrue}
\makeatother

\usepackage{color, colortbl}
\usepackage{bm}
\usepackage{amssymb}
\usepackage{multirow, tabularx}
\usepackage{dsfont}
\usepackage[normalem]{ulem}
\usepackage[pagebackref=true,breaklinks=true,colorlinks,bookmarks=false, allcolors=blue]{hyperref}
\usepackage[capitalize]{cleveref}
\useunder{\uline}{\ul}{}
\usepackage[misc]{ifsym}

\def\eg{\emph{e.g}. } 
\def\ie{\emph{i.e}. } 
 
\def\etc{\emph{etc}. } \def\vs{\emph{vs}. }

\makeatother

\usepackage{xcolor}
\definecolor{COLOR_MEAN}{HTML}{f0f0f0}
\definecolor{ROW_COLOR}{HTML}{C9F7F4}
\definecolor{best}{HTML}{48cae4}
\definecolor{second_best}{HTML}{ffd166}
\definecolor{light_blue}{HTML}{0096c7}
\definecolor{contrast_purple}{HTML}{7b2cbf}
\definecolor{cool_orange}{HTML}{eb5e28}

\newcommand{\gb}{\textbf{GenBench}}
\newcommand*\rot{\rotatebox{90}}
\newcommand{\common}{\includegraphics[height=2.6mm]{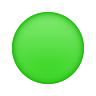}{~Common}}
\newcommand{\finegrained}{\includegraphics[height=2.6mm]{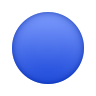}{~Fine-grained}}
\newcommand{\rare}{\includegraphics[height=2.6mm]{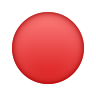}{~Rare}}

\usepackage{xcolor}
\definecolor{COLOR_MEAN}{HTML}{f0f0f0}
\definecolor{ROW_COLOR}{HTML}{C9F7F4}
\definecolor{best}{HTML}{48cae4}
\definecolor{second_best}{HTML}{ffd166}
\definecolor{light_blue}{HTML}{0096c7}
\definecolor{contrast_purple}{HTML}{7b2cbf}
\definecolor{cool_orange}{HTML}{eb5e28}

\begin{document}
%
\title{Benchmarking and Analyzing Generative Data for Visual Recognition}
%
%
%
%

\author{Bo Li, Haotian Liu, Liangyu Chen, Yong Jae Lee \protect\\ Chunyuan Li, Ziwei Liu
\IEEEcompsocitemizethanks{\IEEEcompsocthanksitem Bo Li, Liangyu Chen and Ziwei Liu are with the S-Lab, Nanyang Technological University \protect \\
E-mail: \{libo0013, liangyu.chen, ziwei.liu\}@ntu.edu.sg
\IEEEcompsocthanksitem Haotian Liu and Yong Jae Lee are with University of Wisconsin-Madison.
E-mail: \{lht, yongjaelee\}@cs.wisc.edu
\IEEEcompsocthanksitem Chunyuan Li is with Microsoft Research, Redmond. \protect \\
E-mail: chunyuan.li@microsoft.com}}


%
%

\markboth{Journal of \LaTeX\ Class Files,~Vol.~14, No.~8, August~2015}%
{Shell \MakeLowercase{\textit{et al.}}: Bare Demo of IEEEtran.cls for Computer Society Journals}
%



\IEEEtitleabstractindextext{%
\begin{abstract}
Advancements in large pre-trained generative models have expanded their potential as effective data generators in visual recognition. This work delves into the impact of generative images, primarily comparing paradigms that harness external data (\ie generative \vs retrieval \vs original). Our key contributions are:
\textbf{1) GenBench Construction:} We devise \textbf{GenBench}, a broad benchmark comprising 22 datasets with 2548 categories, to appraise generative data across various visual recognition tasks.
\textbf{2) CLER Score:} To address the insufficient correlation of existing metrics (\eg, FID, CLIP score) with downstream recognition performance, we propose \textbf{CLER}, a training-free metric indicating generative data's efficiency for recognition tasks prior to training.
\textbf{3) New Baselines:} Comparisons of generative data with retrieved data from the same external pool help to elucidate the unique traits of generative data.
\textbf{4) External Knowledge Injection:} By fine-tuning special token embeddings for each category via Textual Inversion, performance improves across 17 datasets, except when dealing with low-resolution reference images. Our exhaustive benchmark and analysis spotlight generative data's promise in visual recognition, while identifying key challenges for future investigation. Our code is available at: \href{https://github.com/Luodian/Genbench}{GenBench}.
\end{abstract}

}

\maketitle

\IEEEdisplaynontitleabstractindextext

%
\IEEEpeerreviewmaketitle
\section{Introduction}
\begin{figure}[htp]
    \centering
    \includegraphics[width=\linewidth]{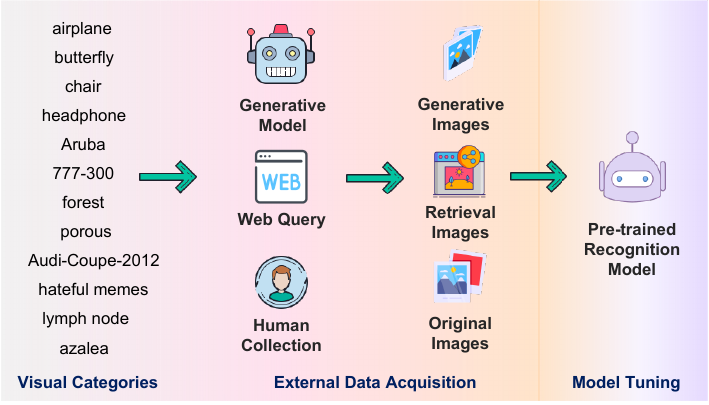}
    \caption{Different external data paradigms for enhancing visual recognition across a wide range of categories.}
    \label{fig:teaser}
\end{figure}

Large-scale pretrained generative models~\cite{rombach2022high,nichol2021glide,ramesh2022hierarchical,saharia2022photorealistic} have advanced the synthesis of realistic images, opening possibilities for improving visual recognition with generative data. Some studies~\cite{he2022synthetic,bansal2023leaving,shipard2023boosting} have explored the use of generative data for training visual systems with positive results. In addition, the development of cost-efficient and controllable generative models~\cite{zhang2023adding,li2023gligen,hu2021lora,chen2022analog,song2020denoising,kong2021fast,blattmann2022retrieval} has gained more and more attention, offering the potential for richer and more cost-effective external data. To determine whether large pre-trained generative models can serve as efficient data generators and to understand the unique characteristics of generative data in comparison to human-annotated and retrieval data (as depicted in \cref{fig:teaser}), a comprehensive study is required.

To address it, we introduce \gb, a benchmark with 22 datasets and 2548 categories, to evaluate the effectiveness of generative data in a broad range of scenarios. Formally, we measure the quality of generative data on \gb by its test accuracy after linear probing on CLIP~\cite{radford2021learning} model and compare it with other types of external data. The datasets on \gb are grouped into three, namely \common, \finegrained, and \rare~concepts, facilitating our analysis of the generative model's generation capabilities on different types of data.

\begin{figure*}[tp]
    \centering
    \includegraphics[width=\linewidth]{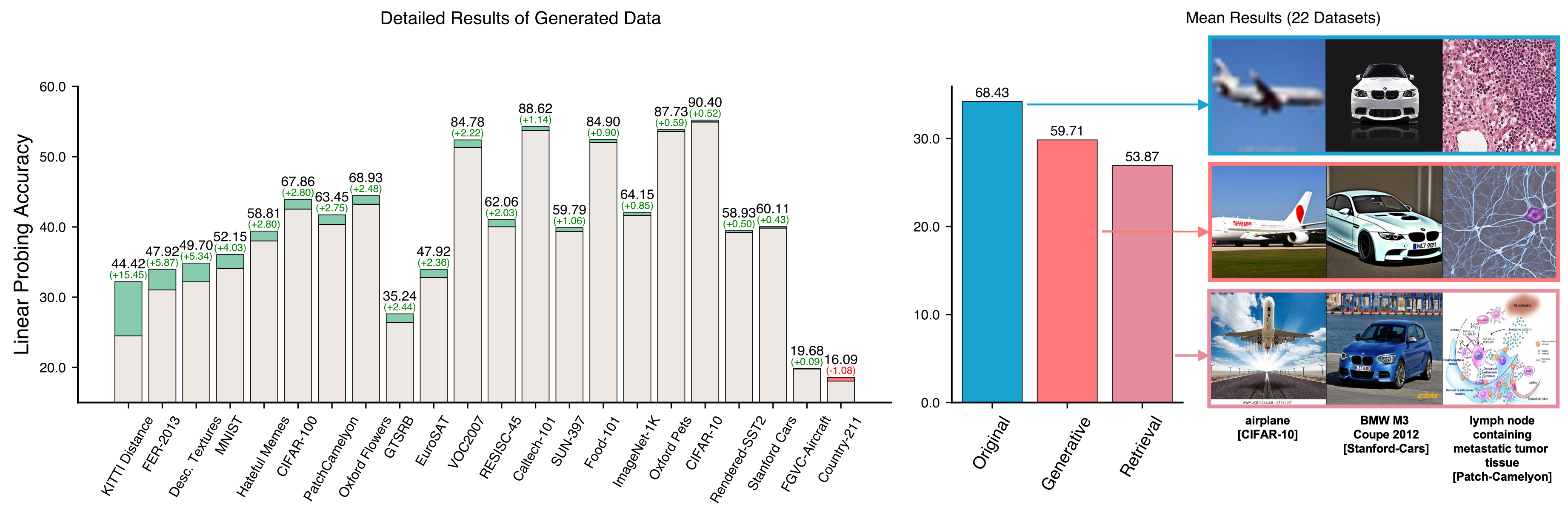}
    \caption{\textbf{Left:} CLIP ViT-B/32 linear probing results for all datasets on \gb, arranged in descending order of improvement over the zero-shot accuracy. The results are based on 500-shot generative data per category with best strategy for each dataset, as shown in \cref{tab:strategy_comparison}. \textbf{Right:} The average results using different external data sources on the 22 datasets, along with sample images for different categories, are shown on the right.}
    \label{fig:rel_improvement}
\end{figure*}

To validate the quality of generative images as well as other types of external data, we propose the Class-centered Recognition (\textbf{CLER}) score. It is directly correlated with the test accuracy from linear probing CLIP~\cite{radford2021learning} and is an efficient training-free measure for assessing the \textit{improvements} of external data.

\begin{figure*}[tp]
    \centering
    \includegraphics[width=\textwidth]{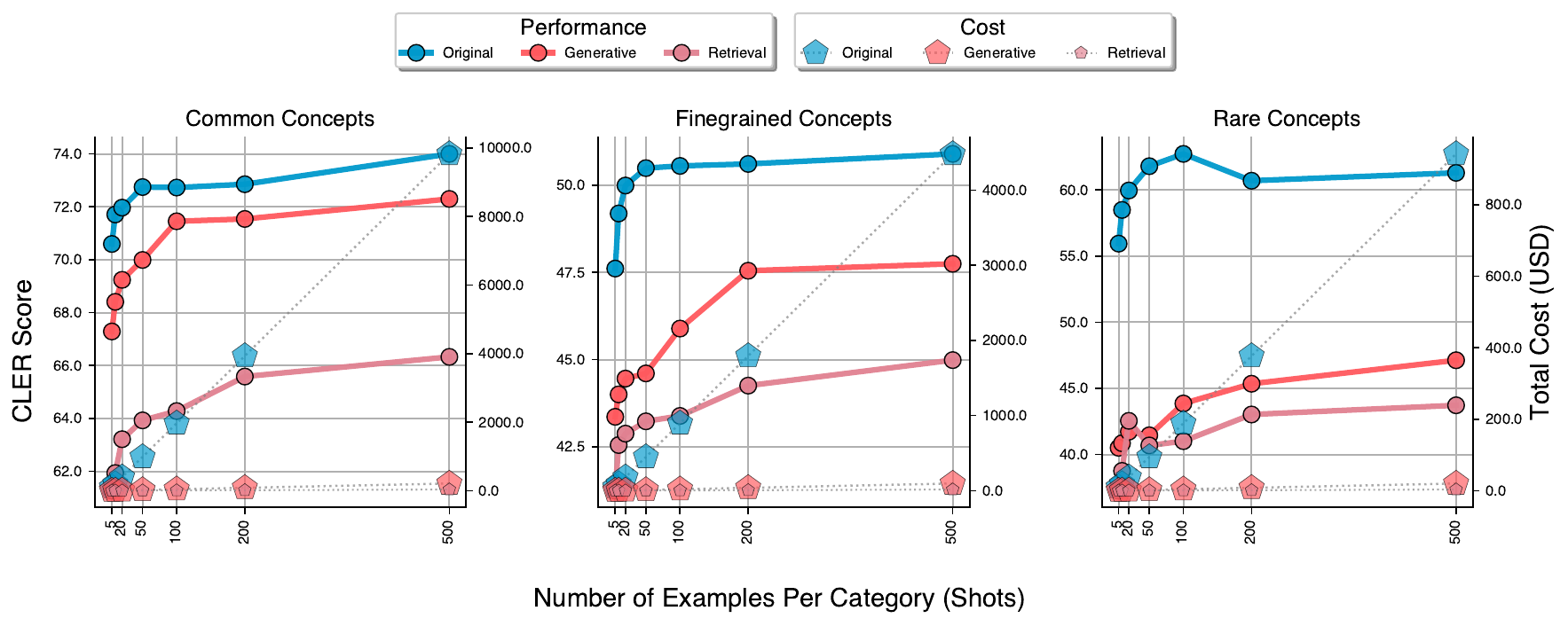}
    \caption{Performance comparison of different types of external data with an increasing amount of data. The cost (in USD) associated with using different amounts of data is labeled on the right.}
    \label{fig:scaling_effect}
\end{figure*}

In~\cref{sec:analysis}, we provide a detailed analysis of generative data. In~\cref{subsec:tradeoffs}, we compare the performance and cost (in USD) of different data types with a maximum of 500 shots per category, totaling over 1M images. Our results in~\cref{fig:scaling_effect} show that generative data is cost-effective and performs well for common concepts. However, it offers no apparent advantage over original data for fine-grained and rare concepts. 


Our integration of~\cref{fig:rel_improvement} and~\cref{fig:scaling_effect} illustrates generative data's superiority over CLIP's zero-shot baseline in varied conceptual realms. ~\cref{fig:rel_improvement} suggests that although generative data scarcely improves CLIP's aptitude for common concepts, it significantly bolsters its handling of nuanced and rare concepts. ~\cref{fig:scaling_effect} emphasizes generative data's cost-effectiveness and advantage over both original and retrieval data. Notwithstanding, limitations surface with rare concepts due to their sparse representation in stable diffusion's training data, impacting the generative data's efficacy.

We then conducted an examination to delineate the possible scenarios for generative data usage and the particular reasons for its infeasibility in certain circumstances. We first delved deeper into the susceptibility of present text-to-image generative models to different prompt strategies, as detailed in ~\cref{subsec:exp_prompt_strategy}. The evidence unearthed highlights the imperative for custom prompt strategies tailored to specific datasets due to the variability of the optimal strategy. Then our investigation in ~\cref{subsec:domain_gap} shines a light on the potential relationship between the amplification effect of generative data and the average text resemblance between the dataset and the pre-training dataset of generative models. We hypothesize the existence of an inherent bias in generative models, which favors categories that either frequently occur or display high query similarity in their pre-training dataset.

Aiming for the augmentation of generative data, we incorporated external knowledge into pre-trained models by fine-tuning specialized token embeddings for each dataset category via Textual Inversion ~\cite{gal2022image}. Upon injecting varied types of reference data, a consistent performance improvement was observed across most of the 17 datasets examined. Despite these advances, certain limitations were observed that the method did not improve generative data performance on specific datasets with only low-resolution reference images.

In summary, our contributions are as follows:
\vspace{-\topsep}
\begin{enumerate}
  \setlength{\parskip}{1pt}
  \setlength{\itemsep}{0pt plus 1pt}
    \item[(1)] A benchmark, \gb, that comprehensively evaluates the benefits of generative data across a diverse range of visual concepts.
    \item[(2)] A analytical training-free metric \textbf{CLER} score for fast evaluation of the image quality of different types of external data.
    \item[(3)] A detailed examination on the potential applications for generative data, along with the explicit rationales for its infeasibility in specified situations.
    \item[(4)] A comprehensive investigation into the effectiveness of generative data through the injection of external knowledge into pretrained generative models.
\end{enumerate}
\vspace{-\topsep}

\section{On the Evaluation of Generative Data}
\label{sec:genbench}
In this section, we investigate how generative data could improve pre-trained visual recognition models on a wider range of downstream datasets, with \gb as our evaluation benchmark.

\subsection{Datasets}
Inherited from existing benchmarks such as Elevater~\cite{li2022elevater} and VTAB~\cite{zhai2019large}, \gb encompasses a total of 22 datasets with a broad spectrum of visual recognition concepts consisting of 2548 categories. These datasets are categorized into the following three main groups. More statistical information on these datasets is listed in the appendix.

\noindent \textbf{\common~Concepts} mainly cover generic objects that are commonly seen in daily life and in the Internet data. In this group, we can evaluate whether the generative models could cover those most common categories effectively. This group includes: ImageNet-1K~\cite{russakovsky2015imagenet}, CIFAR-10~\cite{krizhevsky2009learning}, CIFAR-100~\cite{krizhevsky2009learning}, Caltech-101~\cite{fei2004learning}, VOC-2007~\cite{everingham2009pascal}, MNIST~\cite{lecun1998gradient} and SUN-397~\cite{xiao2010sun}.

\noindent \textbf{\finegrained~Concepts} cover subcategories within a meta-category, such as different breeds of dogs within the dog category or different models of airplanes within the airplane category.
In this group, we can assess the ability of generative models to capture the subtle visual differences between different subcategories and their capacity to generate objects with more specialized names. This task can help identify the limitations of generative models in capturing the complexity and variability of fine-grained object categories. This group includes: Food-101~\cite{bossard2014food}, Oxford Pets~\cite{parkhi2012cats}, Oxford Flowers~\cite{nilsback2008automated}, Standford Cars~\cite{krause20133d}, FGVC Aircraft~\cite{maji2013fine}, Country-211~\cite{radford2021learning}. 

\noindent \textbf{\rare~Concepts} covers less common objects in the real world, such as medical images (e.g., Patch-Camelyon~\cite{veeling2018rotation}) and remote sensing images (e.g., RESISC-45~\cite{cheng2017remote}, EuroSAT~\cite{helber2017eurosat}). Obtaining such data can be challenging, making it important to assess whether generative models can effectively synthesize this type of data, which presents a unique challenge to computer vision research. This group includes PatchCamelyon~\cite{veeling2018rotation}, EuroSAT~\cite{helber2017eurosat}, GTSRB~\cite{stallkamp2011german}, Rendered-SST2~\cite{radford2021learning}, FER 2013~\cite{fer2013}, RESISC-45~\cite{cheng2017remote}, Hateful Memes~\cite{kiela2020hateful}, Describable Textures~\cite{cimpoi2014describing}, and KITTI Distance~\cite{fritsch2013new}.

\subsection{External Data} 
\noindent \textbf{Generative Data.} Although there exists various large-scale pretrained text-to-image generative models~\cite{saharia2022photorealistic,ramesh2022hierarchical}. Our investigation in ~\gb is centered around two readily available and accessible generative models, namely GLIDE ~\cite{nichol2021glide} and Stable Diffusion 2.1~\cite{rombach2022high}. We use the names of categories in the dataset and combine them with various prompt strategies in~\cref{subsec:prompt_strategy} as textual prompts to generative model and obtain corresponding sampled images. If the number of required generated images exceeds the initial prompts, we randomly augment the prompt list to reach the required number.

\noindent \textbf{Retrieval Data.} Similar to retrieval data, we utilize category names along with defined templates specific to each dataset as retrieval queries, without employing additional prompt strategies as used in generative data. We employ text-to-text retrieval on LAION-400M to obtain the top-$k$ results using these queries and select the required number of images based on their similarity to each result. This approach aligns with generative data acquisition to facilitate our evaluation and analysis of data quality. It may differ from other retrieval-based works~\cite{liu2023react}, which we will discuss in \cref{sec: ext_data}. 

\noindent \textbf{Original Data.}  It refers to the training data of each individual dataset. This data type is considered the highest quality among all different types of external data and serves as the upper bound when evaluating on the same amount of shots.
However, it is also the most expensive to acquire.

To compare the cost-effectiveness of different types of data, we also analyze the cost of external data. For generative and retrieval data, it is calculated based on the time needed for sampling generative models or performing top-$k$ text-to-text search and the cost of running a CPU/GPU instance on Azure\footnote{\href{https://instances.vantage.sh/azure/}{Azure Pricing}}. For original data, we provide a reference value for human labeling cost obtained from Amazon Turk\footnote{\href{https://aws.amazon.com/sagemaker/data-labeling/pricing/}{Amazon Turk}}.
We list the estimated cost for each type of external data in ~\cref{tab:data_collection}.


\begin{table}[ht]
\centering
\caption{Data acquisition cost for different types of external data (measured in US dollars per image). See appendix for details.}
\label{tab:data_collection}
\resizebox{\columnwidth}{!}{%
\setlength{\tabcolsep}{10pt}
\renewcommand{\arraystretch}{1.2}
\begin{tabular}{c|c|c}
\toprule
\rowcolor{COLOR_MEAN}
\textbf{Data Type} & \textbf{Collection Source} & \textbf{Est. Cost / Image} \\ \midrule
Generative Data & Model Inference & $2.54 \times 10^{-4}$ USD \\
Retrieval Data & Web/Database Query & $3.93 \times 10^{-5}$ USD \\
Original Data & Human Label & $1.20 \times 10^{-2} $ USD \\ \bottomrule
\end{tabular}%
}
\end{table}



\subsection{Prompt Strategy} 
\label{subsec:prompt_strategy}
Initially, we convert the category name into a textual prompt and feed it into the generative models to produce the corresponding image. The different specific descriptions added to the prompt to generate better images are referred to as prompt strategies. In our benchmark, we use several prompt strategies, which we present below:

\noindent \textbf{Simple Template (ST)} refers the basic prompt format consisting of \textit{a photo of \{\}}, where the category name is inserted into the brackets. 

\noindent \textbf{Defined Template (DT)} refers using simple category names as prompts, we also followed  the practice of CLIP\footnote{\href{https://github.com/openai/CLIP/blob/main/data/prompts.md}{CLIP Prompts}}, by defining dataset-specific prompt formats for each dataset. For example, in FVGC-Aircraft dataset, we extend the simple prompt with a description \textit{a type of aircraft}. 

\noindent \textbf{Category Enhancement (CE)} was introduced in~\cite{he2022synthetic}, where an off-the-shelf word-to-sentence T5~\cite{raffel2020exploring} model is utilized to expand each category into a complete sentence, such as expanding \textit{airplane} into \textit{A large, sleek, white airplane with dual engines is soaring through the clear blue skies, leaving behind a trail of white clouds.}. In the appendix, we will provide examples of generated sentences for different datasets.

\noindent \textbf{Restrictive Description (RD)} are additional, specific phrases added to the prompt in order to guide the generative model to produce images of higher quality. Examples of such phrases include \textit{hi-res, highly detailed, sharp focus}. Additionally, some special restrictive symbols, such as enclosing the category with parentheses, \eg \textit{((airplane))}, can be added to the prompt to make the generative model more focused on the category rather than other descriptive words. These ideas mainly come from the Stable Diffusion community, and the results show that this approach can lead to better generation results, but there is still no rigorous conclusion.

\noindent \textbf{Negative Prompts (NP)} are input arguments that guide the Stable Diffusion model to deliberately exclude particular objects, styles, or abnormalities from the generated image. This significant feature empowers users to eliminate undesired or redundant elements from the final output. Moreover, there are several quality constraints that are applicable to all datasets, such as restricting words like \textit{bad shape} and \textit{misfigured} to prevent the generative model from producing low-quality data.


All prompt strategies aforementioned will be presented with more details for each dataset in the appendix.

\subsection{Evaluation Protocol}

We aim to measure the \textit{improvement} of generative data over the baseline zero-shot accuracy for a diverse range of categories. However, existing metrics such as FID primarily assess the aesthetic quality and visual coherence. While current diffusion-based generative models excel in these metrics, they are inadequate for evaluating generative data's downstream task quality. Typically, the accuracy of a model on test data is used to measure this improvement. Our experiments primarily use CLIP VIT/B-32 as the baseline visual recognition model.

\begin{algorithm}[t]
\caption{Pseudocode of CLER in PyTorch style.}
\label{alg:code}
\algcomment{\fontsize{7.2pt}{0em}\selectfont \texttt{mm}: matrix multiplication; \texttt{T}: transpose.
}
\definecolor{codeblue}{rgb}{0.25,0.5,0.5}
\lstdefinestyle{python}{
  language=python,
  morekeywords={mm,encode_image,encode_text,group,mean,argmax},
  keywordstyle={\fontsize{7.2pt}{7.2pt}\selectfont\textcolor{blue!90!black}}
}
\lstset{
  backgroundcolor=\color{white},
  basicstyle=\fontsize{7.2pt}{7.2pt}\ttfamily\selectfont,
  columns=fullflexible,
  breaklines=true,
  captionpos=b,
  commentstyle=\fontsize{7.2pt}{7.2pt}\color{codeblue},
  mathescape
}
\begin{lstlisting}[style=python]
def CLER_score(gen_images, gen_labels, test_images, test_labels, class_names):
    n_classes = len(class_names)
    gen_emb = CLIP.encode_image(gen_images) $~~$# [M,F]
    test_emb = CLIP.encode_image(test_images) $$# [N,F]
    text_emb = CLIP.encode_text(class_names) $~$# [C,F]
    gen_emb_grp = group(gen_emb, gen_labels) $~$# [C,K,F]
    class_centers = gen_emb_grp.mean(dim=1) $~~$# [C,F]

    preds_CLER = mm(test_emb, class_centers.T)
    preds_CLIP = mm(text_emb, text_features.T)
    preds_ensemble = (preds_CLER + preds_CLIP) / 2

    CLER = sum(preds_CLER.argmax(dim=1) == test_labels)
    CLER_ensemble = sum(preds_ensemble.argmax(dim=1) == test_labels)
\end{lstlisting}
\end{algorithm}

Motivated by CLIP zero-shot evaluation, we propose a training-free metric, Class-centered Recognition (CLER) score, an approximation to the improvement of generative data over a given downstream task. The core idea is to replace the averaged language embeddings of each class in CLIP zero-shot evaluation, with the averaged image embeddings from each class in the target downstream dataset. Specifically, given a set of images $\mathcal{I}$ with corresponding class labels $\mathcal{C}$, a downstream dataset $\mathcal{D}$ containing $C$ classes, CLER score measures the improvement $\mathcal{I}$ can bring to the classification task on $\mathcal{D}$.

First, we extract embeddings of images $\mathcal{I}$ using CLIP $\mathcal{E}_i = \text{CLIP}(\mathcal{I}_i)$.
Then, we group embeddings $\mathcal{E}$ according to labels $\mathcal{T}$, and compute the average embedding within each group: $\mathcal{M}_j = \text{average}(\{\mathcal{E}_i; \mathcal{T}_i = j\})$, where $1~\leq j \leq~C$.
For each test image $\mathcal{D}_k$, we obtain its image embedding $\mathcal{E}_k$ and class prediction $\mathcal{P}_k = \arg\max_{j}(\mathcal{E}_k^\top \cdot \mathcal{E}_j)$.  Then we calculate the mean accuracy of class prediction $\mathcal{P}_k$ for each test image in \cref{eq:cler_score}, where $\mathcal{L}_k$ is the ground truth label of the test image $\mathcal{D}_k$ and $\mathds{1}$ is the indicator function that returns $1$ if $\mathcal{P}_k = \mathcal{L}_k$ or $0$ otherwise.
\begin{equation}
    \text{CLER} = \frac{1}{N} \sum^{1}_{N}\mathds{1}(\mathcal{P}_k = \mathcal{L}_k)
    \label{eq:cler_score}
\end{equation}
 If we specify the CLER score evaluation with the CLIP model,  we can further enhance to a more accurate version, by ensembling the prediction with CLIP contrastive prediction: $\hat{\mathcal{P}}_k = \arg\max_{j}(\mathcal{E}_k^\top \cdot (\mathcal{E}_j + \mathcal{L}_j) / 2)$, where $\mathcal{L}_j$ is the text embeddings obtained from CLIP text encoder for class $j$. We provide the pseudo code for CLER score in ~\cref{alg:code}.
 
 In our experiments and analysis, we primarily use CLIP ViT-B/32 as the recognition model for evaluation. We will also discuss the performance of finetuning the CLIP model on generative data, as well as the evaluation of other backbones (\eg ResNet, larger ViTs, \etc) in the appendix.


\begin{figure}[tp]
    \centering
    \includegraphics[width=\columnwidth]{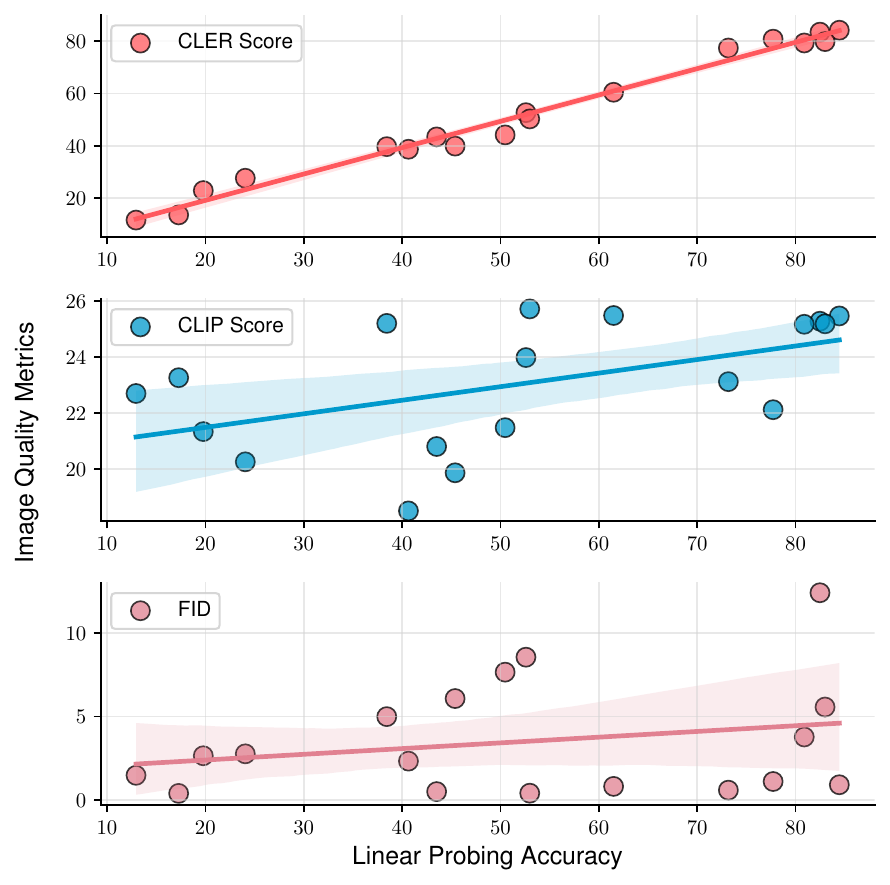}
    \caption{The correlation between three evaluation metrics, CLER score, CLIP score, and FID, with the linear probing accuracy. CLER score demonstrates the best positive correlation among the three metrics.}
    \label{fig:metric_correlation}
\end{figure}

\begin{figure}[tp]
    \centering
    \includegraphics[width=0.9\columnwidth]{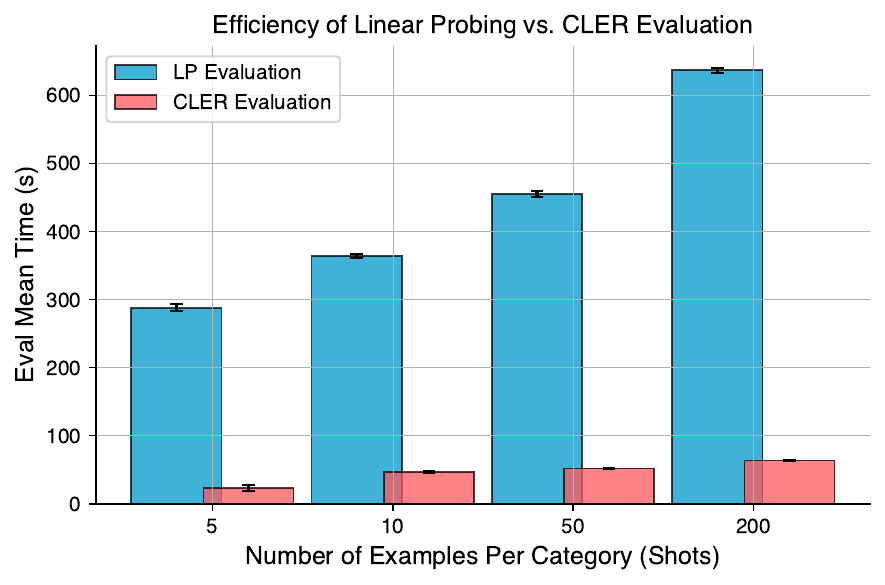}
    \caption{The comparison of evaluation time required for different evaluation metrics used in our experiments.
    We report the average evaluation time with standard errors per dataset for each metric, obtained by running the evaluation three times on a single Nvidia A100 GPU with a dry-run as pre-heat.}
    \label{fig:efficient_evaluation}
\end{figure}

We assess the correlation between CLER score, CLIP score, and FID with linear probing accuracy using a correlation analysis on 18 datasets with 100-shot generative data per category, excluding four datasets with a small number of categories. Results in~\cref{fig:metric_correlation} show that CLER score has the best positive correlation with linear probing accuracy, followed by CLIP score. FID, which is often used to assess image quality, does not exhibit a clear correlation with linear probing accuracy. \textcolor{black}{Moreover, it is important to note that CLER computes dot products between cluster centers, significantly reducing complexity compared to computing dot products for individual images. Consequently, evaluating the CLER score is considerably more cost-effective than obtaining linear probing accuracy, which requires tuning models on generated data, as demonstrated in~\cref{fig:efficient_evaluation}.}

In addition, CLER score can be employed to assess the performance improvement of other types of data, such as original training data or retrieval data from the Internet. In the appendix, we provide a correlation analysis for original and retrieval data. 

Given the advantages of CLER score, we primarily rely on this metric in subsequent analyses, it enables us to efficiently draw indicative conclusions.

\begin{table*}[tp]
\centering
\caption{Results of GLIDE and Stable Diffusion, along with various corresponding strategies, on the \textbf{GenBench} dataset, are presented in the table. We report the CLER scores of the 20-shot generated images on each dataset. For simplicity, we use the following abbreviations: ST for Simple Templates, CE for Category Enhancement, DT for Defined Template, NP for Negative Prompts, RD for Restrictive Descriptions, RA for Retrieval Augmented. We denote the \colorbox{best}{best} and \colorbox{second_best}{second best} with specified colors.}
\label{tab:strategy_comparison}
\resizebox{\textwidth}{!}{%
\setlength{\tabcolsep}{5pt}
\renewcommand{\arraystretch}{2}
\begin{tabular}{c|c|cccccccccccccccccccccc|c}
\toprule
\rowcolor{COLOR_MEAN}
\multicolumn{1}{c|}{\rot{\textbf{Model}}} & \rot{\textbf{Prompt Strategy}} & \rot{\textbf{ImageNet-1K}} & \rot{\textbf{Caltech-101}} & \rot{\textbf{CIFAR-10}} & \rot{\textbf{CIFAR-100}} & \rot{\textbf{Country-211}} & \rot{\textbf{Desc. Textures}} & \rot{\textbf{EuroSAT}} & \rot{\textbf{FER-2013}} & \rot{\textbf{FGVC-Aircraft}} & \rot{\textbf{Food-101}} & \rot{\textbf{GTSRB}} & \rot{\textbf{Hateful Memes}} & \rot{\textbf{KITTI Distance}} & \rot{\textbf{MNIST}} & \rot{\textbf{Oxford Flowers}} & \rot{\textbf{Oxford-IIIT Pets}} & \rot{\textbf{PatchCamelyon}} & \rot{\textbf{Rendered-SST2}} & \rot{\textbf{RESISC-45}} & \rot{\textbf{Stanford Cars}} & \rot{\textbf{SUN-397}} & \rot{\textbf{VOC-2007}} & \rot{\textbf{Mean}} \\ \midrule
\multirow{3}{*}{GLIDE~\cite{nichol2021glide}} & ST & 43.66 & 78.75 & 85.70 & 45.54 & 11.07 & 22.07 & \colorbox{best}{51.36} & \colorbox{second_best}{35.30} & 15.86 & 71.33 & 18.05 & \colorbox{best}{58.61} & 37.27 & 18.98 & 62.23 & 64.24 & \colorbox{best}{62.53} & 52.94 & 49.52 & 51.20 & 49.95 & 80.34 & \textbf{48.48} \\
 & CE & 57.35 & 81.35 & 82.35 & 47.83 & 10.16 & 29.84 & 40.48 & 22.82 & 13.35 & 70.74 & 10.48 & 51.87 & \colorbox{best}{43.32} & 16.29 & 49.15 & 56.89 & \colorbox{second_best}{61.98} & 54.97 & 48.87 & 45.27 & \colorbox{best}{58.57} & 78.62 & \textbf{46.93} \\ 
 & DT & 57.55 & 85.38 & 88.64 & 57.64 & 13.17 & 36.38 & 41.64 & 23.79 & \colorbox{best}{18.96} & 79.58 & 17.75 & \colorbox{second_best}{54.62} & 40.79 & 15.91 & \colorbox{second_best}{65.66} & \colorbox{best}{83.96} & 58.36 & \colorbox{second_best}{55.57} & \colorbox{best}{54.87} & 54.33 & \colorbox{second_best}{58.56}  & 81.73 & \colorbox{second_best}{\textbf{52.04}} \\ \hline
\multirow{6}{*}{Stable Diffusion~\cite{rombach2022high}} & ST & 47.90 & 85.79 & 85.20 & 58.63 & 13.34 & 38.14 & 48.42 & 19.17 & 15.09 & \colorbox{best}{82.39} & 20.99 & 46.05 & 8.30 & 19.80 & 64.87 & 77.41 & 51.72 & 47.83 & 51.31 & 58.35 & 54.35 & 79.20 & \textbf{48.83} \\
 & CE & 60.04 & 81.30 & 88.22 & 56.09 & 11.27 & 30.69 & \colorbox{second_best}{50.00} & 29.81 & 13.24 & 73.95 & 8.73 & 50.67 & 24.47 & 22.15 & 45.95 & 57.01 & 54.70 & 50.58 & 49.15 & 54.11 & 55.90 & 80.97 & \textbf{47.68} \\
 & DT & \colorbox{second_best}{61.07} & 86.31 & 88.90 & 60.19 & 13.85 & \colorbox{best}{46.12} & 40.72 & 27.81 & 16.30 & \colorbox{second_best}{82.29} & \colorbox{best}{30.09} & 49.02 & 9.14 & 19.70 & \colorbox{best}{66.13} & \colorbox{second_best}{83.36} & 53.09 & 50.08 & 52.73 & \colorbox{second_best}{58.45} & 57.48 & \colorbox{best}{82.35} & \textbf{51.60} \\
 & w/ NP & 60.19 & \colorbox{second_best}{86.70} & \colorbox{second_best}{89.43} & \colorbox{second_best}{62.95} & \colorbox{best}{14.25} & 44.36 & 35.70 & 28.11 & 15.95 & 81.64 & 22.03 & 44.75 & 22.08 & 17.49 & 63.31 & 80.87 & 50.54 & 53.27 & 52.34 & 58.41 & 57.55 & \colorbox{second_best}{82.32} & \textbf{51.10} \\
 & w/ NP, RD & \colorbox{best}{61.45} & \colorbox{best}{87.26} & \colorbox{best}{89.22} & \colorbox{best}{64.90} & \colorbox{second_best}{14.14} & \colorbox{second_best}{45.69} & 27.94 & 27.42 & 15.23 & 80.99 & 22.03 & 43.30 & 37.27 & \colorbox{second_best}{26.22} & 64.89 & 80.10 & 53.04 & 55.41 & 50.83 & 57.29 & 57.43 & 82.16 & \textbf{52.01} \\  
\bottomrule
\end{tabular}%
}
\end{table*}
\section{Systematic Analysis on Generative Data}
\label{sec:analysis}
In this section, we analyze: \textbf{(1)} the effectiveness and cost trade-offs of generative data compared to other external data, \textbf{(2)} the performance of different generative models with various prompt strategies on a wide range of categories, and \textbf{(3)} the advantages and disadvantages of using generative data on different datasets and their reasons.

\subsection{Trade-offs between Performance and Cost}
\label{subsec:tradeoffs}
Data efficiency refers to the amount of labeled data required for a model to achieve a desired performance. While more data generally improves recognition accuracy and generalization ability, it also means higher annotation costs. Thus, obtaining higher-quality training data through efficient means is a goal of ML practitioners.

With the information of estimated cost per image in~\cref{tab:data_collection}, in~\cref{fig:scaling_effect}, we demonstrate the impact of adding different amount of external data on performance across different concept groups. This figure provides a detailed breakdown and suggests that there are significant differences in the scaling effect across different concept groups. 

For example, in the \common~concepts, the gap between original and generation is gradually reduced with an increasing amount of data per category. Meanwhile, generative data also significantly outperforms the usage of retrieval data. From the cost perspective, as the amount of generative data increases, the advantage of using generative data over original data becomes more significant in \common~concepts. As the figure suggests, obtaining 500-shot original data on \common~concepts costs around 10k USD while generating the same amount of data would only cost 208 USD. The cost difference between the two is around 9,800 USD, but the performance gap is not significant.

However, in the other two groups, \finegrained~and \rare~concepts, using original data is still advantageous, neither generative data nor retrieval data can reach the performance of original data in these concepts.


\subsection{Performance of Various Prompt Strategies}
\label{subsec:exp_prompt_strategy}
We carried out a comprehensive assessment of the generative data obtained from GLIDE and Stable Diffusion using various prompt strategies on 22 datasets. Table~\ref{tab:strategy_comparison} presents the CLER scores of the 20-shot generative data obtained from different prompt strategies with GLIDE and Stable Diffusion on 22 datasets. 

Our findings indicate that while Stable Diffusion 2.1 with dataset-specific prompts generally yielded better results, the optimal strategy varied across different datasets. Therefore, it is important to consider prompt strategy on a case-by-case basis, as there is no universal optimal strategy. To summarize, our observations are as follows:


First, Stable Diffusion and GLIDE perform comparably well with a simple prompt strategy. Despite Stable Diffusion using a larger pretraining dataset and being considered to create higher-quality images, the ability on creating useful generative data for downstream tasks is not much more significant than GLIDE.

Second, the category enhancement strategy in~\cite{he2022synthetic} may result in performance degradation when evaluated on more datasets due to the introduced noise from expanding a category name into a sentence containing other categories. This approach can enhance data diversity but may lead to images corresponding to multiple concepts but a single label. Compared to GLIDE, Stable Diffusion may be better at handling this kind of noise because it incorporates Open CLIP's Text encoder, which has a better ability to attend to information related to the original category in the sentence.


Third, advanced prompt strategies can further enhance Stable Diffusion's generative ability. Negative prompts and restrictive descriptions both perform well on average. The latter is especially effective in object-centric datasets like CIFAR-10 and ImageNet-1K, where limiting words such as \textit{sharp focus} improve the model's ability to generate object-focused images.

Fourth, the retrieval augmented strategy improves performance for fine-grained and rare concept datasets such as FGVC-Aircraft, Stanford-Cars, and RESISC45. The categories in these datasets are less frequent in the pretraining dataset of the generative models. Retrieval images as image prompt provide more hints for the generative models to generate high-quality data for these rare categories.

Overall, the above analysis mainly stems from an observational perspective. In \cref{subsec:domain_gap}, we will quantitatively analyze the correlation between the CLER score of the generated data and the dataset-level mean text similarity on a specific dataset.

\subsection{Correlation with Domain Gap}
\label{subsec:domain_gap}
From Figure~\ref{fig:rel_improvement} and Figure~\ref{fig:scaling_effect}, it can be seen that the effectiveness of generative data varies across different groups. As these groups were defined based on empirical observations, in this section, we will further quantify the correlation between the performance improvement and the properties of each specific dataset.

We speculate that the capability of a generative model may differ across different categories, which mainly depends on whether the pre-training data it was trained on covers these categories and their related information. We consider measuring the capability of a generative model when facing different categories with {\it mean text similarity} (MTS). 

\begin{figure}[ht]
    \centering
    \includegraphics[width=0.9\columnwidth]{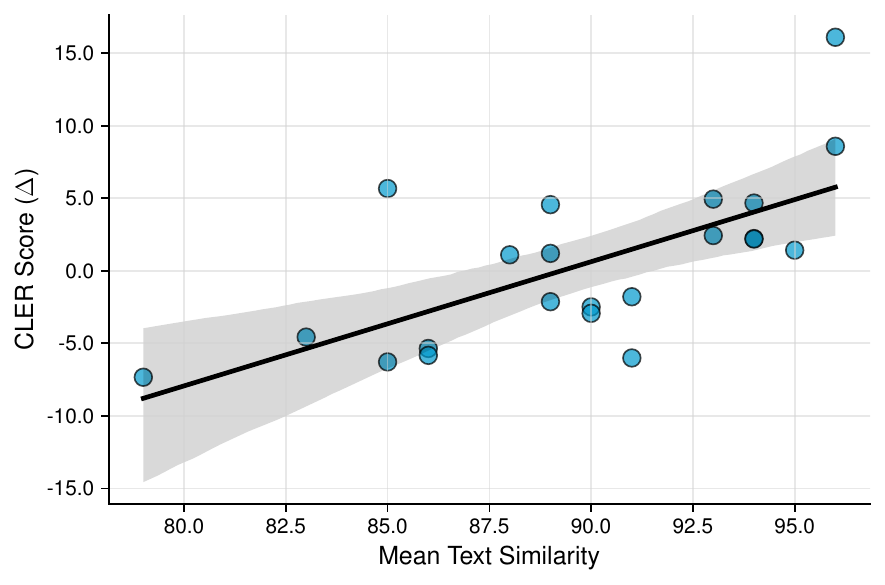}
    \caption{Correlation between CLER score ($\Delta$)  and dataset-level mean text similarity.}
    \label{fig:mean_text_similarity}
\end{figure}

For a specific downstream dataset, we used text-to-text retrieval on LAION-400M, which is a subset of LAION-5B, to retrieve the top-$k$ results using category names as queries. Then, we computed the average similarity among these $k$ retrieval results to obtain the MTS for each category. The category-level MTS measures the similarity between the top-$k$ retrieval results obtained from LAION-400M and the category name query. Common categories such as \textit{airplane} tend to have higher MTS scores, whereas rare categories such as \textit{lymph node} tend to have significantly lower MTS scores. We averaged MTS scores for all categories within a dataset to obtain the dataset-level MTS. From~\cref{fig:mean_text_similarity}, we observed the positive correlation between the changes in CLER score and the dataset-level mean text similarity.

\section{Injecting External Knowledge to Generative Models}
\label{sec:tag}

After observing the improvement that simple data generation can bring to various datasets, we remain interested in exploring whether a more effective method for generating higher quality data can be achieved through efficient fine-tuning using a small amount of data.

Building upon this, we consider injecting (1) retrieval data and (2) original data into generative model using Textual Inversion~\cite{gal2022image}. As shown in the conclusion drawn from \cref{fig:scaling_effect}, the cost of obtaining retrieval data is low, while the cost of acquiring original data is relatively high. Therefore, we consider conducting our experiments using a larger quantity of retrieval data and a limited number of original data.

We adopt the method of Textual Inversion~\cite{gal2022image} to finetune the token embeddings $V*$ (a continous vector representations) for each category using external data.
For example in aeroplane category from VOC-2007 dataset, for retrieval data, we retrieve 100 images related to \textit{aeroplane} from LAION-400M using text-to-text similarity matching. We then use BLIP-2~\cite{li2023blip} to caption these images, and calculate the similarity between the BLIP-2 generated captions and the word \textit{aeroplane} using BERT~\cite{devlin2018bert}. We select the subset of images whose similarity scores are above threshold of 0.8 (the number of remaining images may vary slightly for each category, typically ranging from 5-30 images). The data will be used to train the token embeddings specific to VOC-2007 aeroplane. During training, the conditional text prompt will follow the naming rule of \texttt{\textless|dataset  category|\textgreater} , etc. During sampling, we add defined prompts on those special tokens, forming the conditional sampling prompts \textit{a photo of \texttt{\textless|voc-2007 aeroplane|\textgreater}}.

\begin{figure*}[tp]
    \centering
    \includegraphics[width=1.0\textwidth]{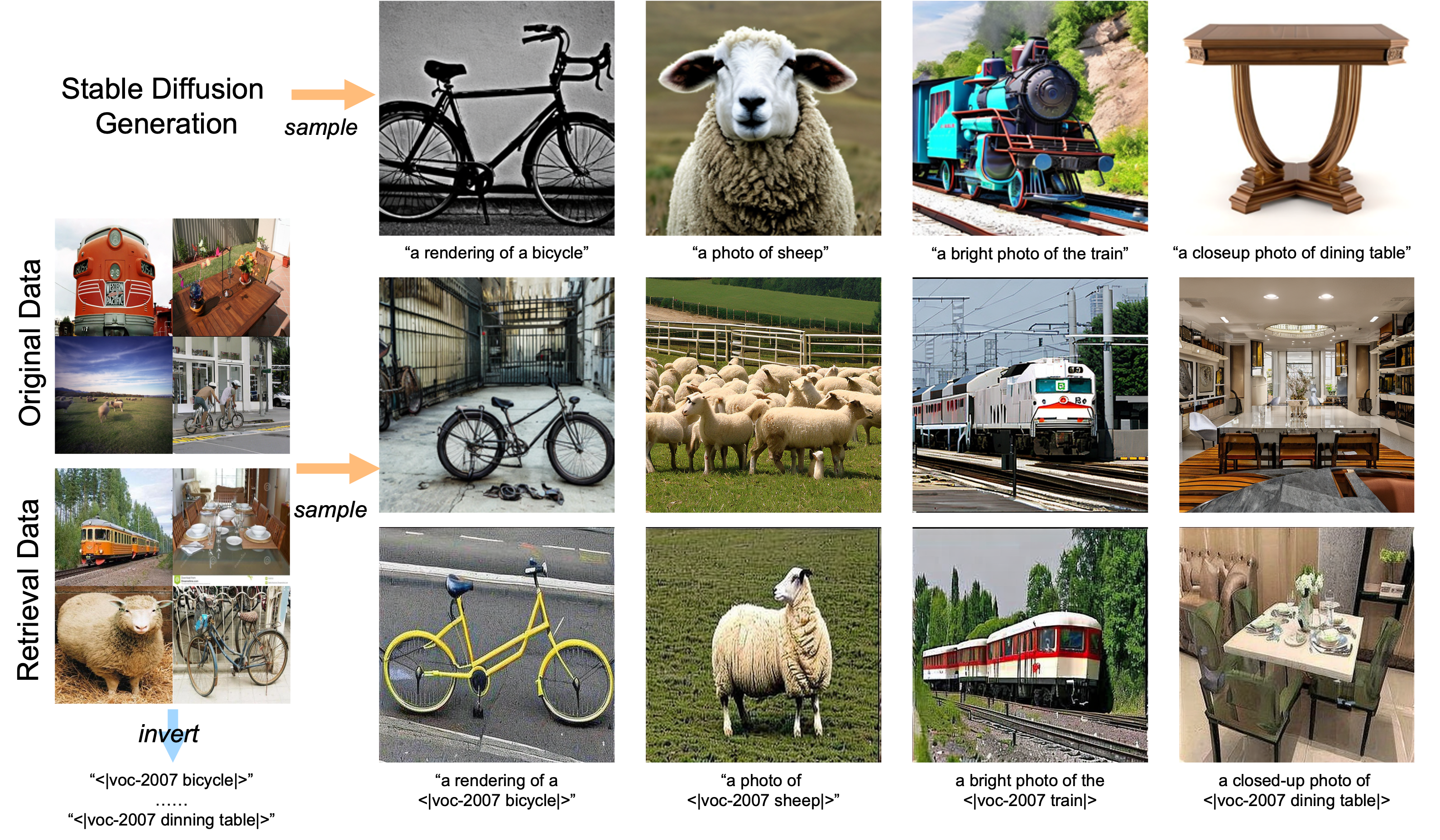}
    \caption{Comparison of the visualization results between direct sampling from Stable Diffusion and finetuned after Textual Inversion~\cite{gal2022image} on original and retrieval data. During textual inversion training, the data are inverted into the form of \texttt{\textless|dataset category|\textgreater}. During sampling, we add diverse prompts on special tokens to obtain different variants.}
    \label{fig:demo}
\end{figure*}

The visual results are provided in \cref{fig:demo}. From a direct comparison between the generation results and Stable Diffusion direct sampling, the results generated after Textual Inversion fine-tuning are closer to the style of the reference data. This approach has the potential to gradually move the sampling space of Stable Diffusion towards the target reference data, thus obtaining data that is more similar to the downstream task.

In \cref{tab:add_external_data}, we evaluated 17 datasets from GenBench. It can be seen that adding retrieval data and original data through Textual Inversion finetuning in Stable Diffusion consistently improves performance across the 17 datasets. We highlighted the column for the datasets where the inclusion of external data further improved the performance. Due to space limit, we provide more detailed analysis in appendix by listing the visual examples of generated images for datasets with significant improvement. We also provide an analysis for why finetuning on original images was not able to improve model performance on low-resolution datasets (\eg for CIFAR-10). These results are presented to provide readers with a reference.

Moreover, Textual Inversion is a very lightweight fine-tuning method. We trained 1000 steps for each category, and the training process takes less than 10 minutes on average per category on a A100 GPU.

\begin{table*}[htp]
\centering
\caption{Analysis of injecting different types of external knowledge by finetuning special token embedding for each category through Textual Inversion across 17 datasets. The \colorbox{best}{entries} indicate improved performance while others indicate degraded performance.}
\label{tab:add_external_data}
\resizebox{\textwidth}{!}{%
\setlength{\tabcolsep}{5pt}
\renewcommand{\arraystretch}{2}
\begin{tabular}{c|cccc|ccccccccccccccccc}
\toprule
\rowcolor{COLOR_MEAN}
\textbf{Model} & \textbf{FT Data} & \textbf{FT Shot} & \textbf{Gen. Shot} & \textbf{Mean} & \rot{\textbf{Caltech-101}} & \rot{\textbf{Country-211}} & \rot{\textbf{Desc. Textures}} & \rot{\textbf{EuroSAT}} & \rot{\textbf{FER-2013}} & \rot{\textbf{FGVC-Aircraft}} & \rot{\textbf{Food-101}} & \rot{\textbf{GTRSB}} & \rot{\textbf{Hateful Memes}} & \rot{\textbf{Kitti Distance}} & \rot{\textbf{Oxford Flowers}} & \rot{\textbf{Oxford Pets}} & \rot{\textbf{PatchCamelyon}} & \rot{\textbf{Rendered-SST2}} & \rot{\textbf{RESISC-45}} & \rot{\textbf{Stanford Cars}} & \rot{\textbf{VOC-2007}} \\ \hline
Stable Diffusion & - & - & 5-shot & 45.92 & 84.68 & \cellcolor[HTML]{48cae4}11.20 & \cellcolor[HTML]{48cae4}40.27 & 37.88 & \cellcolor[HTML]{48cae4}21.04 & \cellcolor[HTML]{48cae4}14.26 & 72.97 & 23.29 & \cellcolor[HTML]{48cae4}44.16 & \cellcolor[HTML]{48cae4}28.69 & \cellcolor[HTML]{48cae4}52.03 & \cellcolor[HTML]{48cae4}72.94 & 51.61 & 49.54 & \cellcolor[HTML]{48cae4}50.48 & \cellcolor[HTML]{48cae4}50.76 & \cellcolor[HTML]{48cae4}74.79 \\ \hline
 & retrieval & 5$\sim$30-shot & 5-shot & 46.46 & 83.70 & \cellcolor[HTML]{48cae4}9.90 & \cellcolor[HTML]{48cae4}36.91 & 23.82 & \cellcolor[HTML]{48cae4}39.78 & \cellcolor[HTML]{48cae4}13.46 & 69.33 & 22.40 & \cellcolor[HTML]{48cae4}49.02 & \cellcolor[HTML]{48cae4}30.80 & \cellcolor[HTML]{48cae4}58.72 & \cellcolor[HTML]{48cae4}74.63 & 50.32 & 50.08 & \cellcolor[HTML]{48cae4}45.64 & \cellcolor[HTML]{48cae4}51.43 & \cellcolor[HTML]{48cae4}79.94 \\ \cline{2-22} 
\multirow{-2}{*}{+ TI} & original & 5-shot & 5-shot & 50.52 & 80.04 & \cellcolor[HTML]{48cae4}14.65 & \cellcolor[HTML]{48cae4}45.05 & 28.56 & \cellcolor[HTML]{48cae4}35.00 & \cellcolor[HTML]{48cae4}17.43 & 69.93 & 16.93 & \cellcolor[HTML]{48cae4}52.10 & \cellcolor[HTML]{48cae4}42.90 & \cellcolor[HTML]{48cae4}71.06 & \cellcolor[HTML]{48cae4}85.50 & 54.45 & 51.13 & \cellcolor[HTML]{48cae4}56.71 & \cellcolor[HTML]{48cae4}57.25 & \cellcolor[HTML]{48cae4}80.16 \\ \hline
Stable Diffusion & - & - & 100-shot & 48.64 & 86.23 & \cellcolor[HTML]{48cae4}12.65 & \cellcolor[HTML]{48cae4}46.38 & 40.04 & \cellcolor[HTML]{48cae4}28.25 & \cellcolor[HTML]{48cae4}13.16 & 74.42 & 26.98 & \cellcolor[HTML]{48cae4}45.20 & \cellcolor[HTML]{48cae4}25.74 & \cellcolor[HTML]{48cae4}57.42 & \cellcolor[HTML]{48cae4}81.82 & 50.27 & 50.80 & \cellcolor[HTML]{48cae4}52.19 & \cellcolor[HTML]{48cae4}55.07 & \cellcolor[HTML]{48cae4}80.29 \\ \hline 
 & retrieval & 5$\sim$30-shot & 100-shot & 47.75 & 85.68 & \cellcolor[HTML]{48cae4}10.79 & \cellcolor[HTML]{48cae4}37.50 & 24.12 & \cellcolor[HTML]{48cae4}46.58 & \cellcolor[HTML]{48cae4}11.88 & 69.24 & 24.45 & \cellcolor[HTML]{48cae4}49.73 & \cellcolor[HTML]{48cae4}32.22 & \cellcolor[HTML]{48cae4}59.42 & \cellcolor[HTML]{48cae4}76.53 & 50.63 & 51.81 & \cellcolor[HTML]{48cae4}47.68 & \cellcolor[HTML]{48cae4}53.21 & \cellcolor[HTML]{48cae4}80.22 \\ \cline{2-22} 
\multirow{-2}{*}{+ TI} & original & 5-shot & 100-shot & 53.90 & 84.40 & \cellcolor[HTML]{48cae4}19.20 & \cellcolor[HTML]{48cae4}48.41 & 34.74 & \cellcolor[HTML]{48cae4}45.86 & \cellcolor[HTML]{48cae4}17.85 & 70.62 & 25.98 & \cellcolor[HTML]{48cae4}55.51 & \cellcolor[HTML]{48cae4}47.65 & \cellcolor[HTML]{48cae4}72.57 & \cellcolor[HTML]{48cae4}84.98 & 55.51 & 52.19 & \cellcolor[HTML]{48cae4}57.85 & \cellcolor[HTML]{48cae4}61.20 & \cellcolor[HTML]{48cae4}81.83 \\ \bottomrule
\end{tabular}%
}
\end{table*}

\section{Related Works}
\subsection{Text-to-Image Generative Models}
 \label{sec: gen_models}
Text-to-image generative models, based on generative adversarial networks \cite{goodfellow2020generative, reed2016generative} or diffusion models \cite{sohl2015deep, ho2020denoising, nichol2021improved}, sample images conditioned on text prompts, \ie to match the given natural language description. 
Several emergent text-to-image models such as Stable Diffusion \cite{rombach2022high}, GLIDE \cite{nichol2021glide}, DALL-E2 \cite{ramesh2022hierarchical}, and Imagen \cite{saharia2022photorealistic} generate high-fiedility synthesized images. Albeit such images achieve perceptual results for human visual communications, their potential utilization for machine visual recognition tasks is yet under-explored. In this paper, we adopt two large-scale pretrained text-to-image generative models, Stable Diffusion and GLIDE, to benchmark and analyze the effect of synthesized images for visual recognition tasks.


\subsection{External Knowledge for Visual Recognition}
 \label{sec: ext_data}

In NLP, several works augment large language models with external knowledge~\cite{peters2019knowledge,guu2020realm,lewis2020retrieval,liu2020k,yu2021dict,borgeaud2021improving,khandelwal2019generalization}.
Motivated by retrieval-augmented models in NLP, several recent works leverage visual and textual knowledge to improve image classification~\cite{long2022retrieval,liu2023react} and multi-modal tasks like question answering~\cite{wu2021multi,marino2021krisp,yang2021empirical,chen2022murag,yasunaga2022retrieval}.
 
A few early attempts at exploring generative data for visual recognition \cite{dosovitskiy2015flownet, peng2017visda, richter2016playing}, usually employ a simulation pipeline with a specific data source, 
 \eg synthetic 2D renderings of 3D models or scenes from graphics engines. Inheriting the limitations of graphics engines, this paradigm usually suffers from the considerable gap between the generated images and real-world observations, as well as the bounded diversity of the specific data sources \cite{he2022synthetic}. Generative models close these gaps with potentially unlimited synthetic data size with a scalable storage cost. \cite{besnier2020dataset} trains image classifiers solely based on the synthesized images generated by a class-conditional BigGAN \cite{brock2018large} model.
\cite{jahanian2021generative} produces multi-view image pairs based on GAN priors to augment contrastive learning.
ImageNet-G-v1 \cite{bansal2023leaving}, a generative dataset based on Stable Diffusion, augments real data to achieve higher accuracy and effective robustness against natural distribution shifts. The most relevant work to ours is \cite{he2022synthetic}, which shows synthetic images are effective in zero-shot/few-shot settings and more performant than ImageNet-1k as pretraining data for downstream tasks. Extending the evaluation of generated images to more domains, we are the first to systematically benchmark their effectiveness along with retrieved and original images. We reveal the failure cases of generated images, and further, propose TIG to improve performance.
\section{Discussions}

\subsection{Takeaway Messages}
In this paper, we present the following key findings:

\noindent \textbf{(1)} Generative data can improve downstream tasks on the most common categories. Obtaining generative data is not significantly more expensive than retrieval data, and it can lead to better performance in downstream tasks.

\noindent \textbf{(2)} The effectiveness of generative data is uncertain for fine-grained and rare categories, and careful selection of prompt strategies is required in these scenarios.

\noindent \textbf{(3)} We found that the effectiveness of generative data is closely related to the mean text similarity (MTS) of downstream tasks. In scenarios with low MTS, using Target-Initialized Generation (TIG) can generate images that are more suitable for downstream tasks.

\noindent \textbf{(4)} Using a few target images as the starting point for generation in TIG can significantly improve the quality of generated images and even outperform the use of original images in some cases.

\subsection{Future Directions and Limitations}
We view generative models as a cost-effective and controllable approach for obtaining high-quality external data. Leveraging retrieval augmentation and other advanced methods, we can further enhance generative models' performance on fine-grained and rare concepts, thereby promoting greater trustworthiness and fairness in downstream tasks. Additionally, our method has the potential to improve long-term learning efficiency, such as providing initial queries for active learning~\cite{chen2022making}.

In our study, we investigated the impact of using up to 500-shot per category (totaling over 1 million images on GenBench), and observed an upward performance trend in~\cref{fig:scaling_effect}. However, exploring the scaling law further with increased computational resources remains a promising direction. Future research should identify key characteristics of generative data and determine optimal scenarios for training downstream models. Increasing prompt diversity within generated images could significantly enhance their utility for model training. Nonetheless, ensuring that such diversity enhancements do not dilute core semantic information poses a challenging yet worthwhile research problem.
\section{Details on Datasets}
In \cref{tab:data_details}, we include the basic statistics of 22 datasets in \textbf{GenBench}. We categorized the dataset into concept groups and listed the number of validation sets and task types for each dataset. 

The majority of the datasets are multi-class tasks, where each image is associated with a single label among multiple categories. VOC 2007 dataset is an exception, as it is a multi-label task, where each image could be associated with multiple labels. Due to the limitations of current generative models and the scope of our study, multi-label task is treated as multi-class task in both for both generative data and retrieval data. Specifically, for each category, we only generate images containing this category and do not consider involving multiple objects.

\begin{table*}[htp]
\centering
\caption{Dataset details, grouped by concepts.}
\label{tab:data_details}
\resizebox{\textwidth}{!}{%
\setlength{\tabcolsep}{12pt}
\renewcommand{\arraystretch}{1.2}
\begin{tabular}{c|c|c|c|c|c}
\toprule
\textbf{Division} & \textbf{Datasets} & \textbf{\#Categories} & \textbf{Validation Size} & \textbf{Evaluation   Metric} & \textbf{Task Type} \\ \midrule
\multirow{7}{*}{\common} & ImageNet-1K & 1,000 & 50,000 & Accuracy & Multi-Class \\
 & CIFAR-10 & 10 & 10,000 & Accuracy & Multi-Class \\
 & CIFAR-100 & 100 & 10,000 & Accuracy & Multi-Class \\
 & Caltech-101 & 101 & 6,085 & Mean Per Class & Multi-Class \\
 & VOC 2007 & 20 & 4,952 & 11-point mAP & Multi-Label \\
 & MNIST & 10 & 10,000 & Accuracy & Multi-Class \\
 & SUN397 & 397 & 19,850 & Accuracy & Multi-Class \\ \midrule
\textbf{Total} & \textbf{-} & \textbf{1,638} & \textbf{110,887} & \textbf{-} & \textbf{-} \\ \midrule
\multirow{6}{*}{\finegrained} & Food-101 & 101 & 25,250 & Accuracy & Multi-Class \\
 & Oxford-IIIT Pets & 37 & 3,669 & Mean Per Class & Multi-Class \\
 & Oxford Flowers & 102 & 6,149 & Mean Per Class & Multi-Class \\
 & Stanford Cars & 196 & 8,041 & Accuracy & Multi-Class \\
 & FGVC Aircraft & 100 & 3,333 & Mean Per Class & Multi-Class \\
 & Country-211 & 211 & 21,100 & Accuracy & Multi-Class \\ \midrule
\textbf{Total} & \textbf{-} & \textbf{747} & \textbf{67,542} & \textbf{-} & \textbf{-} \\ \midrule
\multirow{9}{*}{\rare} & PatchCamelyon & 2 & 32,768 & Accuracy & Multi-Class \\
 & EuroSAT & 10 & 5,000 & Accuracy & Multi-Class \\
 & GTSRB & 43 & 12,630 & Accuracy & Multi-Class \\
 & Rendered-SST2 & 2 & 1,821 & Accuracy & Multi-Class \\
 & FER 2013 & 2 & 3,574 & Accuracy & Multi-Class \\
 & RESISC-45 & 45 & 25,200 & Accuracy & Multi-Class \\
 & Hateful Memes & 2 & 500 & ROC AUC & Multi-Class \\
 & Describable Textures & 47 & 1,880 & Accuracy & Multi-Class \\
 & KITTI Distance & 4 & 711 & Accuracy & Multi-Class \\ \midrule
\textbf{Total} & \textbf{-} & \textbf{157} & \textbf{84,084} & \textbf{-} & \textbf{-} \\ \bottomrule
\end{tabular}%
}
\end{table*}

\section{Details on Measuring Data Aquisation Cost}
In this section, we provide a detailed explanation of how we calculated the acquisition costs for different types of external data.

For generative data, we use the NC24ads A100 v4 instance available on the Azure platform and the Stable Diffusion 2.1 model from~\url{https://github.com/huggingface/diffusers}. The machine contains one A100 GPU, 24 vCPUs, and 220 GB of memory. We estimated the inference time required to generate 1,000 images and calculated the corresponding time and cost per image. 

For retrieval data, since text-to-text search does not require GPU computation, we employed a CPU-based instance, E32ads v5, for this purpose. This machine contains 32 vCPUs and 256 GB of memory, and is less expensive. Due to the fast speed of text-to-text retrieval, we estimated the time required to acquire 10,000 images and scaled it down to the time required for 1,000 images.

To better approximate real-world usage, we used the spot price (in USD) on Azure of both GPU/CPU instance as a reference value. In~\cref{tab:data_acquisation}, we present the prices and the estimated time required to acquire 1,000 images, as well as the calculated cost per image.

It should be noted that the estimated acquisition cost may be affected by the specific GPU and CPU environment, as well as the implementation of model acceleration techniques. In our implementation of the Stable Diffusion inference, we utilized xformers (available at~\url{https://github.com/facebookresearch/xformers}) to optimize memory usage, but did not employ any other quantization strategies.

For original data collected by human labeler, we utilized the reference values provided on~\url{https://aws.amazon.com/cn/sagemaker/data-labeling/pricing/}. It should be noted that the \textit{original data} used in our study refers to the labeled training data provided by the datasets on \textbf{GenBench}, which we consider as equivalent to manually collected human-labeled data.

\begin{table*}[htp]
\centering
\caption{Breakdown on data acquisation cost for different types of external data.}
\label{tab:data_acquisation}
\resizebox{\textwidth}{!}{%
\setlength{\tabcolsep}{12pt}
\renewcommand{\arraystretch}{1.2}
\begin{tabular}{@{}c|c|c|c|c|c@{}}
\toprule
\textbf{Data Type} & \textbf{Collection Source} & \textbf{Instance Type} & \textbf{Price/Hours} & \textbf{Estimate Hours/1K Images} & \textbf{Estimate Cost/Image} \\ \midrule
Generative & Model Inference & NC24ads A100 v4 & 1.47 & 1.74 & 2.54 $\times$ $10^{-4}$ \\
Retrieval & Text to Text Search & E32ads v5 & 0.21 & 0.60 & 3.93 $\times$ $10^{-5}$ \\
Original & Human Labeling & - & - & - & 1.20 $\times 10^{-2}$ \\ \bottomrule
\end{tabular}%
}
\end{table*}

\section{Details on Prompt Strategies}

In~\cref{tab:prompt_examples}, we list the specific prompts for 5 categories under different prompt strategies. Specifically, the categories are: airplane (CIFAR-10), British Shorthair (Oxford-IIT Pets), Acura TL Sedan 2012 (Stanford Cars), and red and white circle no truck entry (GTRSB). The names in brackets refer to corresponding datasets for each category, they also respectively correspond to \common, \finegrained, and \rare~concepts on GenBench.

Specifically, with simple template strategy, only the category name (\eg \textit{airplane}) is given as the text prompt for generative model. With defined template strategy, a more dataset-specific text prompt is created using the CLIP templates~\footnote{\href{https://github.com/openai/CLIP/blob/main/data/prompts.md}{CLIP Templates}}, such as \textit{a black and white photo of the airplane}. In some datasets, there may also be corresponding explanations, such as in FGVC Aircraft dataset, where the template for 737-400 includes the explanation \textit{a type of aircraft}. 

The idea behind the restrictive description strategy mainly comes from suggestions for generating images from the Stable Diffusion community. We consider using limiting words such as \textit{hires} and \textit{sharp focus} and adding \textit{(())} to increase the text encoder's attention weight on these restrictive words. Usually, this type of prompt strategy can help the model generate more object-centric images.

For negative prompts, we consider adding names of other categories within the same dataset during the inference process of Stable Diffusion, in order to restrict the model from generating images of other categories. Empirically, for datasets with a large number of categories, we only consider using a subset of categories (such as randomly selecting 5 categories) as negative prompts.

For category enhancement, the same strategy has been considered in~\cite{he2022synthetic}, we use the same word-to-sentence model to expand a category name into a complete sentence. This approach increases the diversity of text prompts, as well as the final generated images.

We present the images generated using the example prompts listed in Table~\ref{tab:prompt_examples}. By comparing the generated images for each category, the audience can check the differences in image quality generated by different prompts for the same category. This provides a way for evaluating the effectiveness of the different prompt strategies in terms of their ability to produce diverse and high-quality images.

\begin{table*}[htp]
\centering
\caption{The specific prompt example under different prompt strategies.}
\label{tab:prompt_examples}
\resizebox{\textwidth}{!}{%
\setlength{\tabcolsep}{4pt}
\renewcommand{\arraystretch}{1.0}
\begin{tabular}{@{}c|c|c@{}}
\toprule
\textbf{Prompt   Strategy} & \textbf{Example} & \textbf{Category} \\ \midrule
\multirow{5}{*}{Simple Template} & airplane & airplane \\
 & British Shorthair & British Shorthair \\
 & Acura TL Sedan 2012 & Acura TL Sedan 2012 \\
 & 737-400 & 737-400 \\
 & red and white circle no truck entry & red and white circle no turck entry \\ \midrule
\multirow{5}{*}{Defined Template} & a black and white   photo of the airplane. & airplane \\
 & a photo of a British Shorthair, a type of pet. & British Shorthair \\
 & a photo of my old Acura TL Sedan 2012. & Acura TL Sedan 2012 \\
 & a photo of a 737-400, a type of aircraft. & 737-400 \\
 & a close up photo of a red and white circle no truck entry, a traffic   sign. & red and white circle no turck entry \\ \midrule
\multirow{5}{*}{Category Enhancement} & An airplane is sitting on the runway. & airplane \\
 & An all white soldier shows off a shorthaired British & British Shorthair \\
 & a little boy in an Acura TL Sedan in 2012. & Acura TL Sedan 2012 \\
 & a aircraft is reported to be sitting in the plane class for aircraft number 737-400 & 737-400 \\
 & a truck is stamped in red and white during entry & red and white circle no turck entry \\ \midrule
\multirow{5}{*}{Restrictive Description} & a photo of the big   airplane., ((sharp focus)), ((highly detailed)), ((hires)) & airplane \\
 & a photo of a British Shorthair, a type of pet., ((sharp focus)),   ((highly detailed)), ((hires)) & British Shorthair \\
 & a photo of my clean Acura TL Sedan 2012., ((sharp focus)), ((highly   detailed)), ((hires)) & Acura TL Sedan 2012 \\
 & a photo of a 737-400, a type of aircraft., ((sharp focus)), ((highly   detailed)), ((hires)) & 737-400 \\
 & a close up photo of a red and white circle no truck entry, a traffic   sign., ((sharp focus)), ((highly detailed)), ((hires)) & red and white circle no turck entry \\ \midrule
\multirow{5}{*}{Negative Prompts} & \textbf{Prompt:} a good photo   of the airplane. \textbf{Negative Prompt:} automobile, ship, truck, car, forg & airplane \\
 & \textbf{Prompt:} a photo of a British Shorthair, a type of pet. \textbf{Negative Prompt:} Bombay, boxer, leonberger, Persian, Siamese & British Shorthair \\
 & \textbf{Prompt:} a photo of my new Acura TL Sedan 2012. \textbf{Negative Prompt:} Audi V8 Sedan 1994, Bentley Arnage Sedan 2009… & Acura TL Sedan 2012 \\
 & \textbf{Prompt:} a photo of a 737-400, a type of aircraft. \textbf{Negative Prompt:} 737-800, 747-100, 757-200, 767-200, II-76 & 737-400 \\
 & \textbf{Prompt:} a centered photo of a red and white circle no truck entry, a traffic sign. \textbf{Negative Prompt:} empty red and whilte circle & red and white circle no turck entry \\ \bottomrule
\end{tabular}%
}
\end{table*}




\section{Examples on Injecting Different External Data}

In \cref{fig:positive_example1} and \cref{fig:positive_example2}, we select four datasets (FER-2013, Oxford-Flowers, Hateful-Memes, and KITTI-Distance) and provided a comparison of the generated images via direct sampling Stable Diffusion and after Textual Inversion finetuning with different types of external data. These four datasets showed a significant improvement in CLER Score after incorporating external data, as shown in Table 3 from main paper.

For the FER-2013 dataset shown in \cref{fig:positive_example1}, most of the content is related to human faces. The results generated directly by Stable Diffusion for faces were relatively poor (due to privacy concerns, many faces in the training data have varying degrees of mosaic obscuration). However, after incorporating retrieval and original data, the generated images improved significantly for the face region. Among them, the use of original data greatly enhanced the generation of realistic style faces. For the Hateful Memes dataset, it may also be due to a similar reason, where the results generated by Stable Diffusion itself are not good enough.

For fine-grained classification datasets such as Oxford Flowers, incorporating external knowledge can greatly enhance the model's ability to capture category differences and reduce misunderstandings on specific terms. For example, in \cref{fig:positive_example1}, the term \textit{air plant} and \textit{clot foot} in the context of flowers can be confusing because \textit{air}, \textit{plant} and \textit{foot} can be understood separately. This can increase the difficulty for Stable Diffusion to generate the correct category, and thus leading to incorrect results. After introducing external retrieval or original data, we can observe that the generated images become significantly more accurate.

For another type of specific dataset like KITTI Distance, where all data comes from street scenes, we can observe that both the Stable Diffusion direct sampling and incoporating retrieval data can generate decent images but relatively low CLER Score. However, after incorporating original data, the generated images style naturally adapts to the street scenes in the KITTI Distance dataset, which can explain the significant performance improvement on this dataset.

\section{Failures on Finetuning with Low-res Original Data}

\begin{figure*}[htp]
    \centering
    \includegraphics[width=0.9\textwidth]{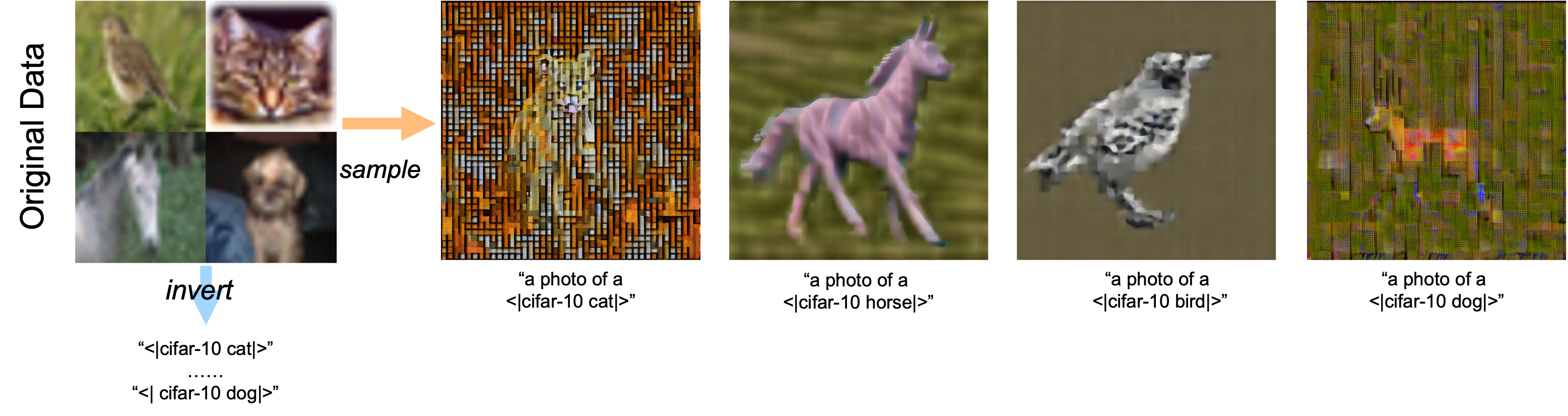}
    \caption{Failure examples on Textual Inversion finetuning with low resolution images.}
    \label{fig:low_res_failures}
\end{figure*}

During finetuning on original data, we also encountered some failure cases~\ref{fig:low_res_failures}. Here, we share them to prevent others from conducting redundant experiments. We found that during Textual Inversion fine-tuning, when the resolution of the reference data is below a certain threshold (less than 32x32 pixels), this subset cannot be generated well by Stable Diffusion. The reconstruction loss during training is difficult to converge, and the generated results, as well as their CLER Scores, are relatively low. This may be because the resolution of the training data used by Stable Diffusion is generally higher than this threshold, resulting in a large domain gap when fine-tuning on low-resolution images. In Figure 2, we share a failure case on the CIFAR-10 dataset.

\section{Extended Analysis}
In order to demonstrate that the CLER Score can effectively verify data quality and quantify the improvement of downstream tasks after adding the data to training, we conducted correlation analysis on linear probing accuracy in the main paper. In this section, we further investigate the CLER score performance when evaluating on different architectures.


\begin{figure*}[htp]
    \centering
    \includegraphics[width=\linewidth]{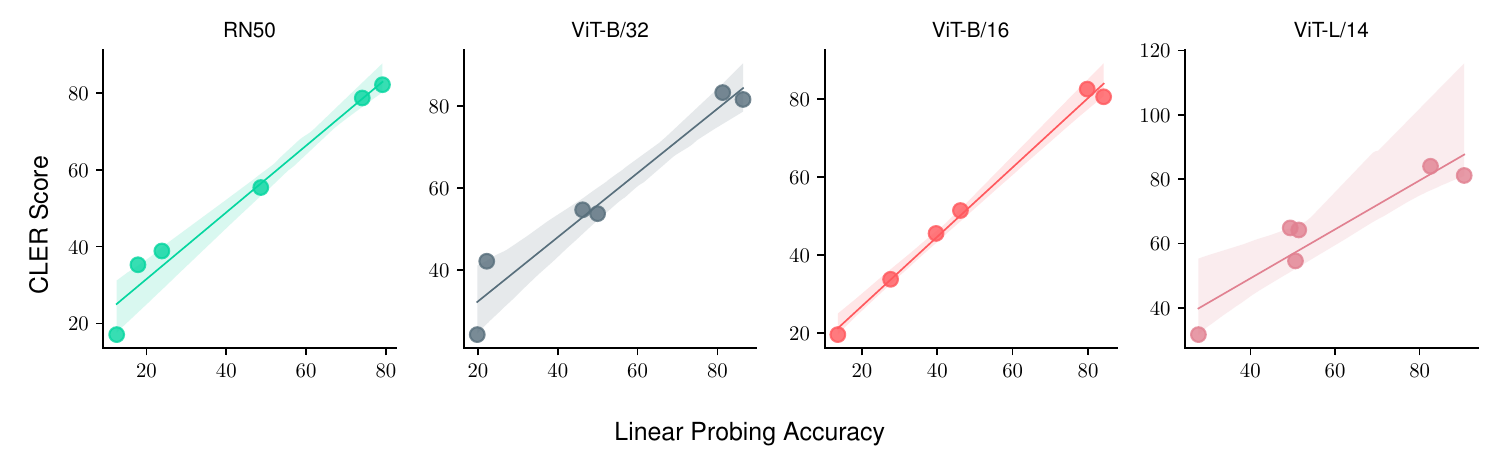}
    \caption{Correlation between CLER score and linear probing accuracy on 5-shot generative data with different architectures.}
    \label{fig:arch_scaling}
\end{figure*}

\subsection{Correlation with Other External Data}

\begin{figure}[htp]
    \centering
    \includegraphics[width=0.85\columnwidth]{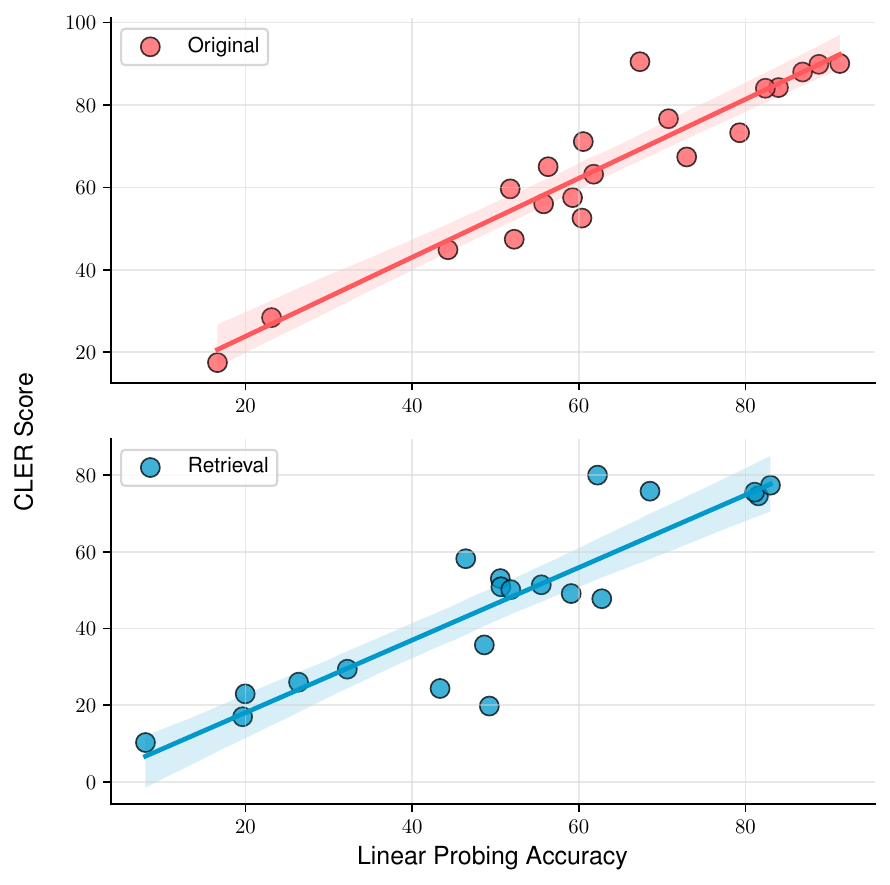}
    \caption{Correlation between CLER score and linear probing accuracy with 5-shot original and retrieval data.}
    \label{fig:other_external}
\end{figure}

In~\cref{fig:other_external}, we examined the correlation between CLER score and linear probe accuracy on other types of external data, including both original and retrieval data. As depicted in the figure, we observed a strong correlation between CLER score and linear probe accuracy for the original data, while the correlation was slightly weaker for the retrieval data. These results indicate that CLER score can be applied more broadly to assess the quality of external data, including both human-labeled and machine-generated data. Moreover, the efficient validation of external data quality provided by CLER score can be particularly beneficial for downstream tasks that require large amounts of labeled data, as it can reduce the time and cost of collecting and labeling data. Additionally, since CLER score does not require any training, it can be used as a quick and easy method to evaluate the quality of external data before incorporating it into downstream tasks. However, it should be noted that while CLER score is strongly correlated with linear probe accuracy, further research is still needed to fully explore its ability to represent finetuned accuracy on downstream tasks.

\subsection{Correlation with Retrieval Metrics}

From a conceptual standpoint, the CLER Score essentially quantifies how closely generated images align with the per-class cluster centers of downstream test data within the embedding space. A smaller distance leads to a higher CLER score, thereby enabling generative models to optimize further and produce high-quality images beneficial for downstream tasks.

Moreover, we acknowledge that the image-to-image retrieval precision-recall metric shares a similar concepts. Thus, we conduct experiments on ImageNet-1k datasets, utilizing generated images from various generative models and configurations. For each dataset, we performed image-to-image retrieval per class by computing CLIP embeddings of both generated and test images, retrieving the top-5 nearest test embeddings based on cosine similarity, and calculating precision@5 and recall@5 metrics. Subsequently, we conducted a correlation analysis between retrieval precision and the CLER Score. These additional experiments and discussions have been incorporated into Section 5.1 (highlighted in blue), providing a comparative analysis between CLER and retrieval precision-recall metrics to further validate the effectiveness of the CLER metric.

\subsection{Evaluation on Other Architectures}

To evaluate performance on different architectures, we selected several other pre-trained models provided by CLIP, including ResNet-50, VIT-B/32, VIT-B/16, and VIT-L/14, ranked by the number of model parameters. Due to computational resource constraints, the experiments are conducted on six datasets (Caltech-101, EuroSAT, FGVC Aircraft, Hateful Memes, MNIST, and VOC-2007). From~\cref{fig:arch_scaling}, as observed in the results, a strong correlation was observed between CLER score and linear probing accuracy for all the pre-trained models, even when the model parameters were increased. Although the ViT-L/14 model has relatively larger errors on some datasets, its CLER score and linear probing accuracy remained highly correlated. However, for larger model parameters (\eg exceeding 1 billion parameters), it remains unclear whether CLER score can continue to indicatively evaluate the quality of external data. Further investigation is needed to address this question.

\begin{figure*}[htp]
    \centering
    \includegraphics[width=\textwidth]{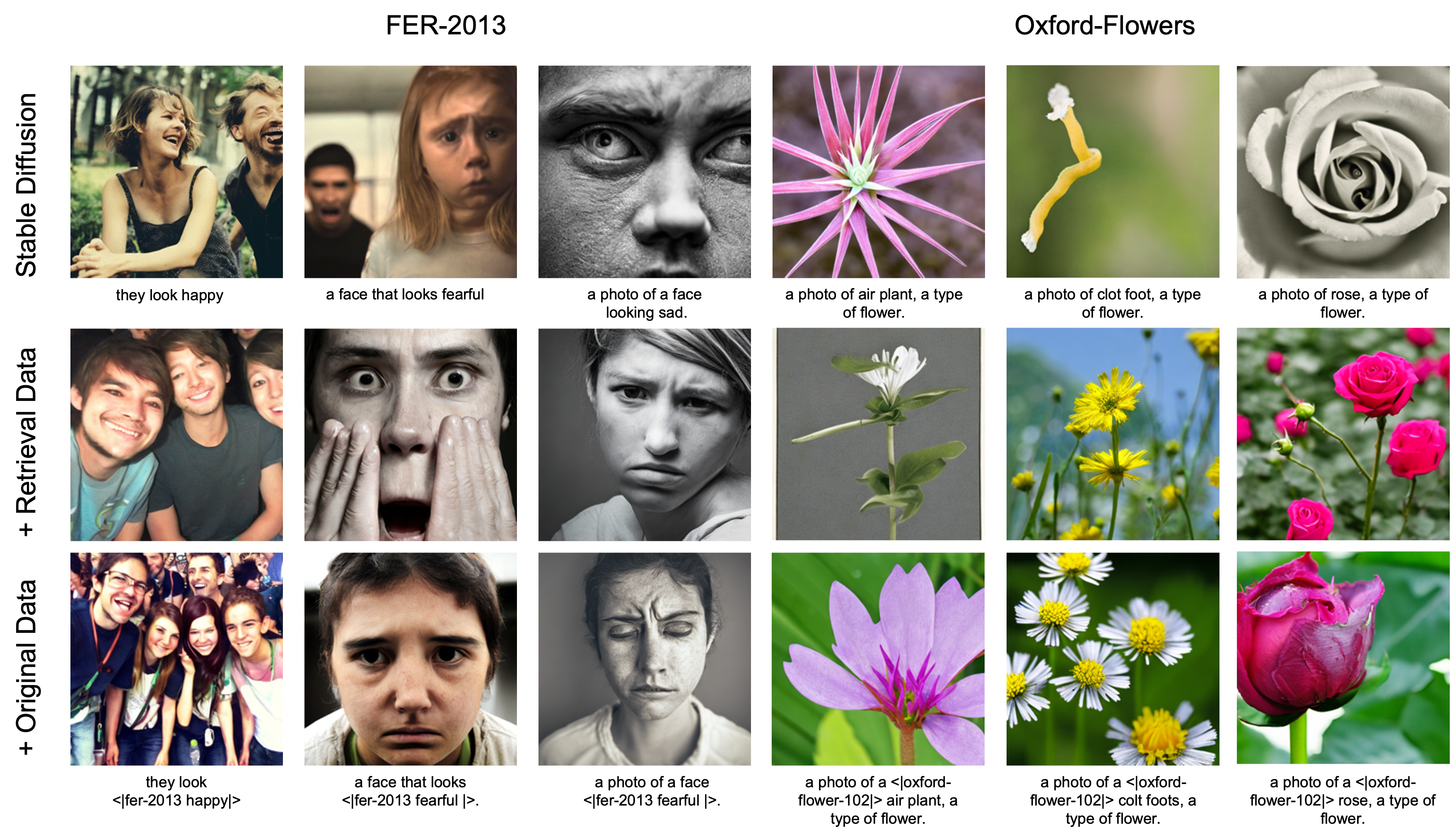}
    \caption{The first row represents the examples generated via directly sampling Stable Diffusion, while the second and third rows represent the results generated using Textual Inversion fine-tuning with different external data combinations.}
    \label{fig:positive_example1}
\end{figure*}

\begin{figure*}[htp]
    \centering
    \includegraphics[width=\textwidth]{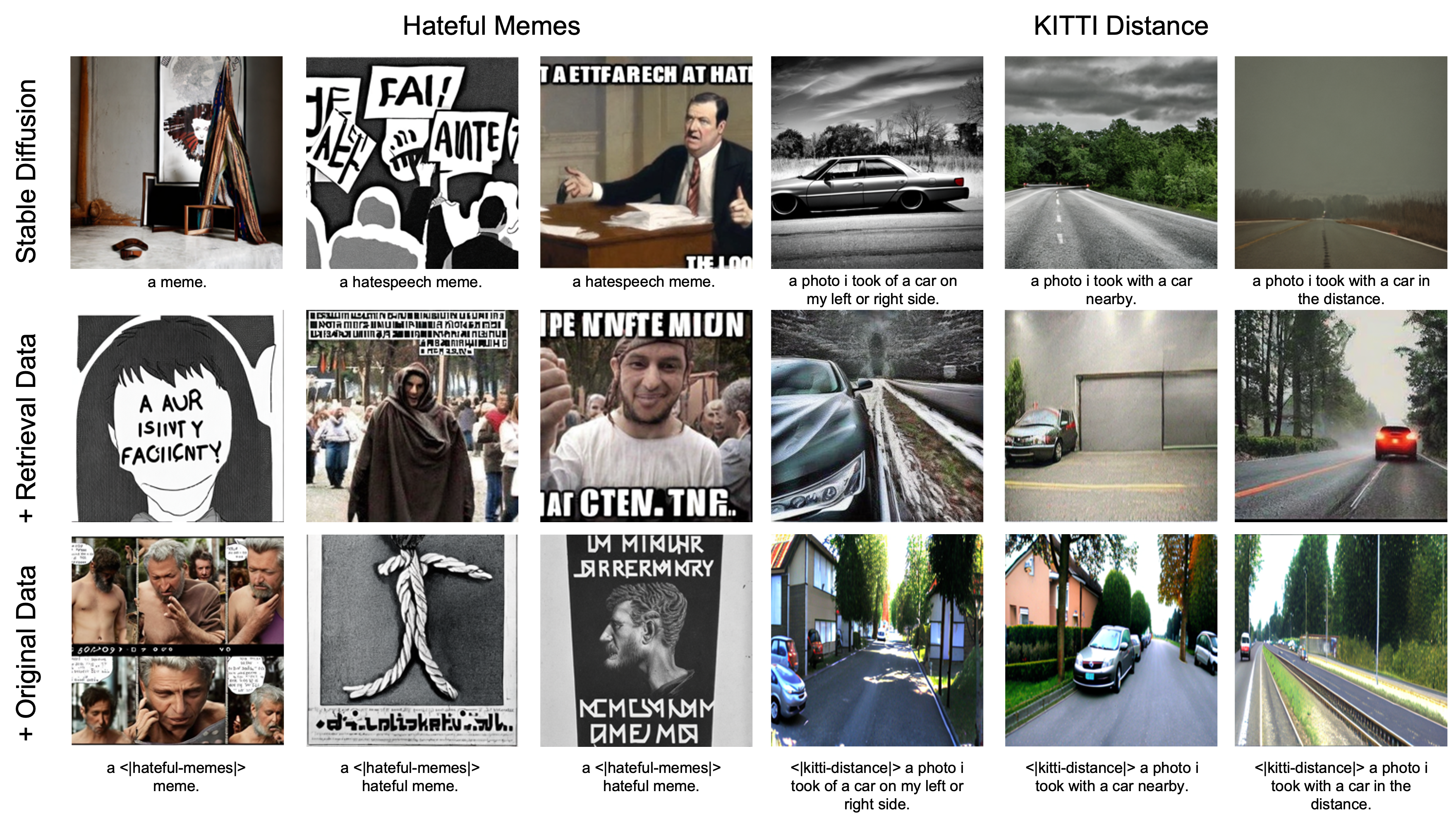}
    \caption{The first row represents the examples generated via directly sampling Stable Diffusion, while the second and third rows represent the results generated using Textual Inversion fine-tuning with different external data combinations.}
    \label{fig:positive_example2}
\end{figure*}

\subsection{Discussion on Distilling Pretrained Knowledge into Downstream Models}

\textcolor{black}{Indeed, we acknowledge that our approach primarily focuses on distilling pretrained image generation models into downstream recognition models. Given the rapid proliferation of pretrained models trained extensively on large-scale datasets with substantial computational resources, our method explicitly leverages this prevailing trend. Specifically, we outline how our approach effectively capitalizes on the rich knowledge embedded in these larger generative models, highlighting practical trade-offs and efficiency improvements for downstream recognition tasks.}

\subsection{Discussion on Limitations on Rare Concepts}

\textcolor{black}{As mentioned above, we also face limitations when pretrained generation models do not have enough knowledge about rare concepts. These limitations mainly appear as poor performance when generating images from rare categories, leading to lower effectiveness in downstream tasks. However, we consider this issue temporary, as generative models will likely improve with larger training datasets and more computational resources. Therefore, our method of distilling knowledge from these models remains effective and useful despite current limitations. We have also clearly explained dataset biases and the specific difficulties involved in rare and fine-grained categories. Moreover, we highlight that using external knowledge injection methods like Textual Inversion can help reduce these biases, providing a practical way to overcome current challenges and improve future model performance.}

\section*{Acknowledgement}
This study is supported by the Ministry of Education, Singapore, under its MOE AcRF Tier 2 (MOE-T2EP20221-0012, MOE-T2EP20223-0002), and under the RIE2020 Industry Alignment Fund – Industry Collaboration Projects (IAF-ICP) Funding Initiative, as well as cash and in-kind contribution from the industry partner(s).

\begin{IEEEbiography}[{\includegraphics[width=1in,height=1.25in,clip]{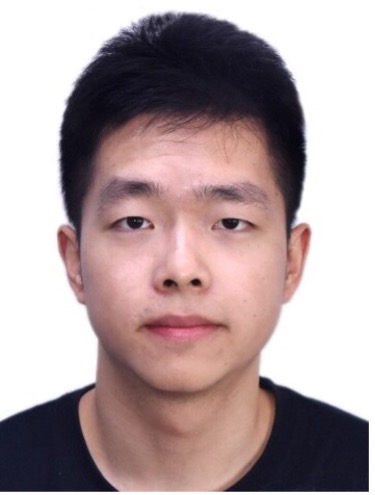}}]{Bo Li} received the B.S. from Harbin Institute of Technology in 2020. He is currently working towards the PhD degree in the College of Computer and Data Science, Nanyang Technological University, Singapore. His research interests mainly include multimodal learning and foundation models.
\end{IEEEbiography}

\begin{IEEEbiography}[{\includegraphics[width=1in,height=1.25in,clip]{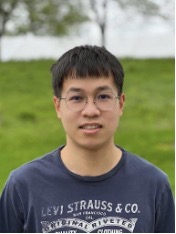}}]{Haotian Liu} received the B.E. (with honors) from Zhejiang University and the Ph.D. degree from the University of Wisconsin–Madison in 2024, under the supervision of Prof. Yong Jae Lee. He is currently a member of technical staff at xAI. His research interests include computer vision, multimodal learning, and foundation models.
\end{IEEEbiography}

\begin{IEEEbiography}[{\includegraphics[width=1in,height=1.25in,clip,keepaspectratio]{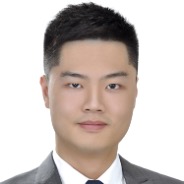}}]{Liangyu Chen} received his B.Eng. from Nanyang Technological University, Singapore, in 2022. He worked at MMLab@NTU from 2022 to 2024. He is currently pursuing a Ph.D. in Computer Science at Stanford University. He researches multimodal foundation models and data-centric machine learning.
\end{IEEEbiography}

\begin{IEEEbiography}[{\includegraphics[width=1in,height=1.25in,clip,keepaspectratio]{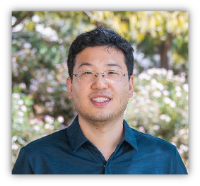}}]{Yong Jae Lee} is an Associate Professor in the Department of Computer Sciences at the University of Wisconsin-Madison. His main research interests are in computer vision and machine learning. He is particularly interested in creating robust AI systems that can understand our multimodal world with minimal human supervision. He received his Ph.D. from UT-Austin, and was a postdoc at CMU and UC Berkeley.
\end{IEEEbiography}

\begin{IEEEbiography}[{\includegraphics[width=1in,height=1.25in,clip,keepaspectratio]{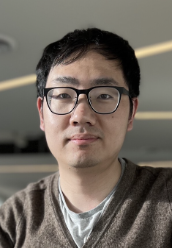}}]{Chunyuan Li}’s recent focus is large multimodal models in vision-and-language. His contributions include the development of LLaVA and the series of model families, as well as earlier works such as Oscar, GLIP, Grounding DINO, GLIGEN and Florence. He has worked with xAI, ByteDance, Microsoft Research, and obtained his PhD at Duke University.

\end{IEEEbiography}

\begin{IEEEbiography}[{\includegraphics[width=1in,height=1.25in,clip,keepaspectratio]{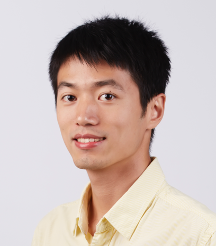}}]{Ziwei Liu} is currently an Associate Professor at Nanyang Technological University, Singapore. His research revolves around computer vision, machine learning and computer graphics. He has published extensively on top-tier conferences and journals in relevant fields, including CVPR, ICCV, ECCV, NeurIPS, ICLR, SIGGRAPH, TPAMI, TOG and Nature - Machine Intelligence. He is the recipient of PAMI Mark Everingham Prize, CVPR Best Paper Award Candidate, Asian Young Scientist Fellowship, International Congress of Basic Science Frontiers of Science Award and MIT Technology Review Innovators under 35 Asia Pacific. He serves as an Area Chair of CVPR, ICCV, ECCV, NeurIPS and ICLR, as well as an Associate Editor of IJCV.
\end{IEEEbiography}

\ifCLASSOPTIONcaptionsoff
  \newpage
\fi



%
{\small
\bibliographystyle{plain}
\bibliography{egbib}

\begin{thebibliography}{10}

\bibitem{fer2013}
{FER} 2013: Kaggle challenges in representation learning facial expression
  recognition.
\newblock \url{https://www.kaggle.com/}.

\bibitem{bansal2023leaving}
Hritik Bansal and Aditya Grover.
\newblock Leaving reality to imagination: Robust classification via generated
  datasets.
\newblock {\em arXiv preprint arXiv:2302.02503}, 2023.

\bibitem{besnier2020dataset}
Victor Besnier, Himalaya Jain, Andrei Bursuc, Matthieu Cord, and Patrick
  P{\'e}rez.
\newblock This dataset does not exist: training models from generated images.
\newblock In {\em ICASSP 2020-2020 IEEE International Conference on Acoustics,
  Speech and Signal Processing (ICASSP)}, pages 1--5. IEEE, 2020.

\bibitem{blattmann2022retrieval}
Andreas Blattmann, Robin Rombach, Kaan Oktay, and Bj{\"o}rn Ommer.
\newblock Retrieval-augmented diffusion models.
\newblock {\em arXiv preprint arXiv:2204.11824}, 2022.

\bibitem{borgeaud2021improving}
Sebastian Borgeaud, Arthur Mensch, Jordan Hoffmann, Trevor Cai, Eliza
  Rutherford, Katie Millican, George van~den Driessche, Jean-Baptiste Lespiau,
  Bogdan Damoc, Aidan Clark, et~al.
\newblock Improving language models by retrieving from trillions of tokens.
\newblock {\em arXiv preprint arXiv:2112.04426}, 2021.

\bibitem{bossard2014food}
Lukas Bossard, Matthieu Guillaumin, and Luc~Van Gool.
\newblock Food-101--mining discriminative components with random forests.
\newblock In {\em ECCV}, 2014.

\bibitem{brock2018large}
Andrew Brock, Jeff Donahue, and Karen Simonyan.
\newblock Large scale gan training for high fidelity natural image synthesis.
\newblock {\em arXiv preprint arXiv:1809.11096}, 2018.

\bibitem{chen2022making}
Liangyu Chen, Yutong Bai, Siyu Huang, Yongyi Lu, Bihan Wen, Alan~L Yuille, and
  Zongwei Zhou.
\newblock Making your first choice: To address cold start problem in vision
  active learning.
\newblock {\em arXiv preprint arXiv:2210.02442}, 2022.

\bibitem{chen2022analog}
Ting Chen, Ruixiang Zhang, and Geoffrey Hinton.
\newblock Analog bits: Generating discrete data using diffusion models with
  self-conditioning.
\newblock {\em arXiv preprint arXiv:2208.04202}, 2022.

\bibitem{chen2022murag}
Wenhu Chen, Hexiang Hu, Xi~Chen, Pat Verga, and William~W Cohen.
\newblock Murag: Multimodal retrieval-augmented generator for open question
  answering over images and text.
\newblock {\em arXiv preprint arXiv:2210.02928}, 2022.

\bibitem{cheng2017remote}
Gong Cheng, Junwei Han, and Xiaoqiang Lu.
\newblock Remote sensing image scene classification: Benchmark and state of the
  art.
\newblock {\em Proceedings of the IEEE}, 2017.

\bibitem{cimpoi2014describing}
Mircea Cimpoi, Subhransu Maji, Iasonas Kokkinos, Sammy Mohamed, and Andrea
  Vedaldi.
\newblock Describing textures in the wild.
\newblock In {\em CVPR}, 2014.

\bibitem{devlin2018bert}
Jacob Devlin, Ming-Wei Chang, Kenton Lee, and Kristina Toutanova.
\newblock Bert: Pre-training of deep bidirectional transformers for language
  understanding.
\newblock {\em arXiv preprint arXiv:1810.04805}, 2018.

\bibitem{dosovitskiy2015flownet}
Alexey Dosovitskiy, Philipp Fischer, Eddy Ilg, Philip Hausser, Caner Hazirbas,
  Vladimir Golkov, Patrick Van Der~Smagt, Daniel Cremers, and Thomas Brox.
\newblock Flownet: Learning optical flow with convolutional networks.
\newblock In {\em Proceedings of the IEEE international conference on computer
  vision}, pages 2758--2766, 2015.

\bibitem{everingham2009pascal}
Mark Everingham, Luc Van~Gool, Christopher~KI Williams, John Winn, and Andrew
  Zisserman.
\newblock The pascal visual object classes (voc) challenge.
\newblock {\em International journal of computer vision}, 88:303--308, 2009.

\bibitem{fei2004learning}
Li~Fei-Fei, Rob Fergus, and Pietro Perona.
\newblock Learning generative visual models from few training examples: An
  incremental bayesian approach tested on 101 object categories.
\newblock In {\em 2004 conference on computer vision and pattern recognition
  workshop}, pages 178--178. IEEE, 2004.

\bibitem{fritsch2013new}
Jannik Fritsch, Tobias Kuehnl, and Andreas Geiger.
\newblock A new performance measure and evaluation benchmark for road detection
  algorithms.
\newblock In {\em ITSC}. IEEE, 2013.

\bibitem{gal2022image}
Rinon Gal, Yuval Alaluf, Yuval Atzmon, Or~Patashnik, Amit~H Bermano, Gal
  Chechik, and Daniel Cohen-Or.
\newblock An image is worth one word: Personalizing text-to-image generation
  using textual inversion.
\newblock {\em arXiv preprint arXiv:2208.01618}, 2022.

\bibitem{goodfellow2020generative}
Ian Goodfellow, Jean Pouget-Abadie, Mehdi Mirza, Bing Xu, David Warde-Farley,
  Sherjil Ozair, Aaron Courville, and Yoshua Bengio.
\newblock Generative adversarial networks.
\newblock {\em Communications of the ACM}, 63(11):139--144, 2020.

\bibitem{guu2020realm}
Kelvin Guu, Kenton Lee, Zora Tung, Panupong Pasupat, and Ming-Wei Chang.
\newblock Realm: Retrieval-augmented language model pre-training.
\newblock {\em arXiv preprint arXiv:2002.08909}, 2020.

\bibitem{he2022synthetic}
Ruifei He, Shuyang Sun, Xin Yu, Chuhui Xue, Wenqing Zhang, Philip Torr, Song
  Bai, and Xiaojuan Qi.
\newblock Is synthetic data from generative models ready for image recognition?
\newblock {\em arXiv preprint arXiv:2210.07574}, 2022.

\bibitem{helber2017eurosat}
Patrick Helber, Benjamin Bischke, Andreas Dengel, and Damian Borth.
\newblock Eurosat: A novel dataset and deep learning benchmark for land use and
  land cover classification, 2017.

\bibitem{ho2020denoising}
Jonathan Ho, Ajay Jain, and Pieter Abbeel.
\newblock Denoising diffusion probabilistic models.
\newblock {\em Advances in Neural Information Processing Systems},
  33:6840--6851, 2020.

\bibitem{hu2021lora}
Edward~J Hu, Yelong Shen, Phillip Wallis, Zeyuan Allen-Zhu, Yuanzhi Li, Shean
  Wang, Lu~Wang, and Weizhu Chen.
\newblock Lora: Low-rank adaptation of large language models.
\newblock {\em arXiv preprint arXiv:2106.09685}, 2021.

\bibitem{jahanian2021generative}
Ali Jahanian, Xavier Puig, Yonglong Tian, and Phillip Isola.
\newblock Generative models as a data source for multiview representation
  learning.
\newblock {\em arXiv preprint arXiv:2106.05258}, 2021.

\bibitem{khandelwal2019generalization}
Urvashi Khandelwal, Omer Levy, Dan Jurafsky, Luke Zettlemoyer, and Mike Lewis.
\newblock Generalization through memorization: Nearest neighbor language
  models.
\newblock {\em arXiv preprint arXiv:1911.00172}, 2019.

\bibitem{kiela2020hateful}
Douwe Kiela, Hamed Firooz, Aravind Mohan, Vedanuj Goswami, Amanpreet Singh,
  Pratik Ringshia, and Davide Testuggine.
\newblock The hateful memes challenge: Detecting hate speech in multimodal
  memes.
\newblock {\em NeurIPS}, 2020.

\bibitem{kong2021fast}
Zhifeng Kong and Wei Ping.
\newblock On fast sampling of diffusion probabilistic models.
\newblock {\em arXiv preprint arXiv:2106.00132}, 2021.

\bibitem{krause20133d}
Jonathan Krause, Michael Stark, Jia Deng, and Li~Fei-Fei.
\newblock 3d object representations for fine-grained categorization.
\newblock In {\em ICCV workshops}, 2013.

\bibitem{krizhevsky2009learning}
Alex Krizhevsky, Geoffrey Hinton, et~al.
\newblock Learning multiple layers of features from tiny images.
\newblock 2009.

\bibitem{lecun1998gradient}
Yann LeCun, L{\'e}on Bottou, Yoshua Bengio, and Patrick Haffner.
\newblock Gradient-based learning applied to document recognition.
\newblock {\em Proceedings of the IEEE}, 86(11):2278--2324, 1998.

\bibitem{lewis2020retrieval}
Patrick Lewis, Ethan Perez, Aleksandra Piktus, Fabio Petroni, Vladimir
  Karpukhin, Naman Goyal, Heinrich K{\"u}ttler, Mike Lewis, Wen-tau Yih, Tim
  Rockt{\"a}schel, et~al.
\newblock Retrieval-augmented generation for knowledge-intensive {NLP} tasks.
\newblock {\em NeurIPS}, 2020.

\bibitem{li2022elevater}
Chunyuan Li, Haotian Liu, Liunian~Harold Li, Pengchuan Zhang, Jyoti Aneja,
  Jianwei Yang, Ping Jin, Yong~Jae Lee, Houdong Hu, Zicheng Liu, et~al.
\newblock Elevater: A benchmark and toolkit for evaluating language-augmented
  visual models.
\newblock {\em arXiv preprint arXiv:2204.08790}, 2022.

\bibitem{li2023blip}
Junnan Li, Dongxu Li, Silvio Savarese, and Steven Hoi.
\newblock Blip-2: Bootstrapping language-image pre-training with frozen image
  encoders and large language models.
\newblock {\em arXiv preprint arXiv:2301.12597}, 2023.

\bibitem{li2023gligen}
Yuheng Li, Haotian Liu, Qingyang Wu, Fangzhou Mu, Jianwei Yang, Jianfeng Gao,
  Chunyuan Li, and Yong~Jae Lee.
\newblock Gligen: Open-set grounded text-to-image generation.
\newblock {\em arXiv preprint arXiv:2301.07093}, 2023.

\bibitem{liu2023react}
Haotian Liu, Kilho Son, Jianwei Yang, Ce~Liu, Jianfeng Gao, Yong~Jae Lee, and
  Chunyuan Li.
\newblock Learning customized visual models with retrieval-augmented knowledge.
\newblock 2023.

\bibitem{liu2020k}
Weijie Liu, Peng Zhou, Zhe Zhao, Zhiruo Wang, Qi~Ju, Haotang Deng, and Ping
  Wang.
\newblock K-{BERT}: Enabling language representation with knowledge graph.
\newblock In {\em AAAI}, 2020.

\bibitem{long2022retrieval}
Alexander Long, Wei Yin, Thalaiyasingam Ajanthan, Vu~Nguyen, Pulak Purkait,
  Ravi Garg, Alan Blair, Chunhua Shen, and Anton van~den Hengel.
\newblock Retrieval augmented classification for long-tail visual recognition.
\newblock In {\em Proceedings of the IEEE/CVF Conference on Computer Vision and
  Pattern Recognition}, pages 6959--6969, 2022.

\bibitem{maji2013fine}
Subhransu Maji, Esa Rahtu, Juho Kannala, Matthew Blaschko, and Andrea Vedaldi.
\newblock Fine-grained visual classification of aircraft.
\newblock {\em arXiv preprint arXiv:1306.5151}, 2013.

\bibitem{marino2021krisp}
Kenneth Marino, Xinlei Chen, Devi Parikh, Abhinav Gupta, and Marcus Rohrbach.
\newblock Krisp: Integrating implicit and symbolic knowledge for open-domain
  knowledge-based {VQA}.
\newblock In {\em CVPR}, 2021.

\bibitem{nichol2021glide}
Alex Nichol, Prafulla Dhariwal, Aditya Ramesh, Pranav Shyam, Pamela Mishkin,
  Bob McGrew, Ilya Sutskever, and Mark Chen.
\newblock Glide: Towards photorealistic image generation and editing with
  text-guided diffusion models.
\newblock {\em arXiv preprint arXiv:2112.10741}, 2021.

\bibitem{nichol2021improved}
Alexander~Quinn Nichol and Prafulla Dhariwal.
\newblock Improved denoising diffusion probabilistic models.
\newblock In {\em International Conference on Machine Learning}, pages
  8162--8171. PMLR, 2021.

\bibitem{nilsback2008automated}
Maria-Elena Nilsback and Andrew Zisserman.
\newblock Automated flower classification over a large number of classes.
\newblock In {\em Indian Conference on Computer Vision, Graphics \& Image
  Processing}. IEEE, 2008.

\bibitem{parkhi2012cats}
Omkar~M Parkhi, Andrea Vedaldi, Andrew Zisserman, and CV~Jawahar.
\newblock Cats and dogs.
\newblock In {\em CVPR}, 2012.

\bibitem{peng2017visda}
Xingchao Peng, Ben Usman, Neela Kaushik, Judy Hoffman, Dequan Wang, and Kate
  Saenko.
\newblock Visda: The visual domain adaptation challenge.
\newblock {\em arXiv preprint arXiv:1710.06924}, 2017.

\bibitem{peters2019knowledge}
Matthew~E Peters, Mark Neumann, Robert~L Logan~IV, Roy Schwartz, Vidur Joshi,
  Sameer Singh, and Noah~A Smith.
\newblock Knowledge enhanced contextual word representations.
\newblock {\em arXiv preprint arXiv:1909.04164}, 2019.

\bibitem{radford2021learning}
Alec Radford, Jong~Wook Kim, Chris Hallacy, Aditya Ramesh, Gabriel Goh,
  Sandhini Agarwal, Girish Sastry, Amanda Askell, Pamela Mishkin, Jack Clark,
  et~al.
\newblock Learning transferable visual models from natural language
  supervision.
\newblock In {\em ICML}, 2021.

\bibitem{raffel2020exploring}
Colin Raffel, Noam Shazeer, Adam Roberts, Katherine Lee, Sharan Narang, Michael
  Matena, Yanqi Zhou, Wei Li, and Peter~J Liu.
\newblock Exploring the limits of transfer learning with a unified text-to-text
  transformer.
\newblock {\em The Journal of Machine Learning Research}, 21(1):5485--5551,
  2020.

\bibitem{ramesh2022hierarchical}
Aditya Ramesh, Prafulla Dhariwal, Alex Nichol, Casey Chu, and Mark Chen.
\newblock Hierarchical text-conditional image generation with clip latents.
\newblock {\em arXiv preprint arXiv:2204.06125}, 2022.

\bibitem{reed2016generative}
Scott Reed, Zeynep Akata, Xinchen Yan, Lajanugen Logeswaran, Bernt Schiele, and
  Honglak Lee.
\newblock Generative adversarial text to image synthesis.
\newblock In {\em International conference on machine learning}, pages
  1060--1069. PMLR, 2016.

\bibitem{richter2016playing}
Stephan~R Richter, Vibhav Vineet, Stefan Roth, and Vladlen Koltun.
\newblock Playing for data: Ground truth from computer games.
\newblock In {\em Computer Vision--ECCV 2016: 14th European Conference,
  Amsterdam, The Netherlands, October 11-14, 2016, Proceedings, Part II 14},
  pages 102--118. Springer, 2016.

\bibitem{rombach2022high}
Robin Rombach, Andreas Blattmann, Dominik Lorenz, Patrick Esser, and Bj{\"o}rn
  Ommer.
\newblock High-resolution image synthesis with latent diffusion models.
\newblock In {\em Proceedings of the IEEE/CVF Conference on Computer Vision and
  Pattern Recognition}, pages 10684--10695, 2022.

\bibitem{russakovsky2015imagenet}
Olga Russakovsky, Jia Deng, Hao Su, Jonathan Krause, Sanjeev Satheesh, Sean Ma,
  Zhiheng Huang, Andrej Karpathy, Aditya Khosla, Michael Bernstein, et~al.
\newblock Imagenet large scale visual recognition challenge.
\newblock {\em International journal of computer vision}, 115:211--252, 2015.

\bibitem{saharia2022photorealistic}
Chitwan Saharia, William Chan, Saurabh Saxena, Lala Li, Jay Whang, Emily
  Denton, Seyed Kamyar~Seyed Ghasemipour, Burcu~Karagol Ayan, S~Sara Mahdavi,
  Rapha~Gontijo Lopes, et~al.
\newblock Photorealistic text-to-image diffusion models with deep language
  understanding.
\newblock {\em arXiv preprint arXiv:2205.11487}, 2022.

\bibitem{shipard2023boosting}
Jordan Shipard, Arnold Wiliem, Kien~Nguyen Thanh, Wei Xiang, and Clinton
  Fookes.
\newblock Boosting zero-shot classification with synthetic data diversity via
  stable diffusion.
\newblock {\em arXiv preprint arXiv:2302.03298}, 2023.

\bibitem{sohl2015deep}
Jascha Sohl-Dickstein, Eric Weiss, Niru Maheswaranathan, and Surya Ganguli.
\newblock Deep unsupervised learning using nonequilibrium thermodynamics.
\newblock In {\em International Conference on Machine Learning}, pages
  2256--2265. PMLR, 2015.

\bibitem{song2020denoising}
Jiaming Song, Chenlin Meng, and Stefano Ermon.
\newblock Denoising diffusion implicit models.
\newblock {\em arXiv preprint arXiv:2010.02502}, 2020.

\bibitem{stallkamp2011german}
Johannes Stallkamp, Marc Schlipsing, Jan Salmen, and Christian Igel.
\newblock The german traffic sign recognition benchmark: a multi-class
  classification competition.
\newblock In {\em IJCNN}, 2011.

\bibitem{veeling2018rotation}
Bastiaan~S Veeling, Jasper Linmans, Jim Winkens, Taco Cohen, and Max Welling.
\newblock Rotation equivariant cnns for digital pathology.
\newblock In {\em MICCAI}, 2018.

\bibitem{wu2021multi}
Jialin Wu, Jiasen Lu, Ashish Sabharwal, and Roozbeh Mottaghi.
\newblock Multi-modal answer validation for knowledge-based {VQA}.
\newblock {\em arXiv preprint arXiv:2103.12248}, 2021.

\bibitem{xiao2010sun}
Jianxiong Xiao, James Hays, Krista~A Ehinger, Aude Oliva, and Antonio Torralba.
\newblock Sun database: Large-scale scene recognition from abbey to zoo.
\newblock In {\em 2010 IEEE computer society conference on computer vision and
  pattern recognition}, pages 3485--3492. IEEE, 2010.

\bibitem{yang2021empirical}
Zhengyuan Yang, Zhe Gan, Jianfeng Wang, Xiaowei Hu, Yumao Lu, Zicheng Liu, and
  Lijuan Wang.
\newblock An empirical study of {GPT}-3 for few-shot knowledge-based {VQA}.
\newblock {\em arXiv preprint arXiv:2109.05014}, 2021.

\bibitem{yasunaga2022retrieval}
Michihiro Yasunaga, Armen Aghajanyan, Weijia Shi, Rich James, Jure Leskovec,
  Percy Liang, Mike Lewis, Luke Zettlemoyer, and Wen-tau Yih.
\newblock Retrieval-augmented multimodal language modeling.
\newblock {\em arXiv preprint arXiv:2211.12561}, 2022.

\bibitem{yu2021dict}
Wenhao Yu, Chenguang Zhu, Yuwei Fang, Donghan Yu, Shuohang Wang, Yichong Xu,
  Michael Zeng, and Meng Jiang.
\newblock Dict-bert: Enhancing language model pre-training with dictionary.
\newblock {\em arXiv preprint arXiv:2110.06490}, 2021.

\bibitem{zhai2019large}
Xiaohua Zhai, Joan Puigcerver, Alexander Kolesnikov, Pierre Ruyssen, Carlos
  Riquelme, Mario Lucic, Josip Djolonga, Andre~Susano Pinto, Maxim Neumann,
  Alexey Dosovitskiy, et~al.
\newblock A large-scale study of representation learning with the visual task
  adaptation benchmark.
\newblock {\em arXiv preprint arXiv:1910.04867}, 2019.

\bibitem{zhang2023adding}
Lvmin Zhang and Maneesh Agrawala.
\newblock Adding conditional control to text-to-image diffusion models.
\newblock {\em arXiv preprint arXiv:2302.05543}, 2023.

\end{thebibliography}
}

%








\end{document}